%% file: manuscript.tex
\documentclass[lettersize,journal]{IEEEtran}
\usepackage{amsmath,amsfonts}
\usepackage{algorithmic}
\usepackage{algorithm}
\usepackage{array}
\usepackage[caption=false,font=normalsize,labelfont=sf,textfont=sf]{subfig}
\usepackage{textcomp}
\usepackage{stfloats}
\usepackage{url}
\usepackage{verbatim}
\usepackage{graphicx}
\usepackage{cite}
\usepackage{multirow, multicol, makecell} 
\usepackage{booktabs}
\usepackage{soul}
\usepackage[table,dvipsnames]{xcolor}
\usepackage{bbding, pifont, utfsym}
\newcommand{\TableOuterRule}{\Xhline{1.3pt}}
\newcommand{\TableInnerRule}{\Xhline{1.0pt}}
\newcommand{\SotaAltRowColor}{gray!18}
\newcommand{\SotaOursRowColor}{blue!12}
\newcommand{\SotaAltRow}{\rowcolor{\SotaAltRowColor}}
\newcommand{\SotaOursRow}{\rowcolor{\SotaOursRowColor}}

\begin{document}

\title{LCPNet: Latent Consistent Proximal Unfolding Network\\ for Infrared Small Target Detection}

\author{Tianfang Zhang, Lei Li, Chang Liu, Zhenming Peng, Huaping Zhang, Xiangyang Ji
\thanks{Corresponding author: \textit{Xiangyang~Ji}.}
\thanks{Tianfang Zhang, Chang Liu and Xiangyang Ji are with the Department of Automation, Tsinghua University, Beijing, 100190, China. (E-mail: sparkcarleton@gmail.com; \{liuchang2022, xyji\}@tsinghua.edu.cn)}
\thanks{Zhenming Peng are with the School of Information and Communication Engineering and the Laboratory of Imaging Detection and Intelligent Perception, University of Electronic Science and Technology of China, Chengdu, 610054, China. (E-mail: zmpeng@uestc.edu.cn)}
\thanks{Lei Li and Huaping Zhang are with the School of Artificial Intelligence, Beijing Institute of Technology, Beijing, 100081, China. (E-mail: \{lilei, kevinzhang\}@bit.edu.cn)}

}

\markboth{Journal of \LaTeX\ Class Files,~Vol.~14, No.~8, August~2021}%
{Shell \MakeLowercase{\textit{et al.}}: A Sample Article Using IEEEtran.cls for IEEE Journals}


\maketitle

\begin{abstract}
Infrared small target detection (IRSTD) aims to identify long distance small targets from complex infrared backgrounds, and is a fundamental task in remote sensing. 
Deep learning methods have improved IRSTD by learning discriminative image-to-mask mappings, but such feed-forward designs often underuse physical decomposition structure between targets and backgrounds. 
Deep unfolding methods partially address this issue by embedding model-driven iterations into neural networks, yet existing designs still operate mainly in image domain and use updates and memory mechanisms that are not fully coupled with underlying optimization process. 
To address these limitations, we propose Latent Consistent Proximal unfolding network (LCPNet). 
First, we verify that low-rank prior remains valid in latent representations and perform unfolding in this space, preserving physical constraint while avoiding repeated compression of intermediate states. 
Second, we derive a Latent Consistent Proximal (LCP) solver that evolves each latent variable from its previous state rather than reconstructing through an indirect residual, and stabilizes small target updates through task-adaptive normalization and gain control. 
Third, we introduce Shared Optimization Memory (SOM), a common historical state shared by all decomposition variables to provide coordinated guidance across unfolding stages. 
Extensive experiments on four public benchmarks demonstrate that LCPNet achieves accurate and robust detection with low false-alarm rates, together with competitive efficiency among high-accuracy deep unfolding methods.
Model and code are available at \url{https://github.com/Tianfang-Zhang/LCPNet}.
\end{abstract}


\section{Introduction}
\IEEEPARstart{I}{nfrared} small target detection (IRSTD) is a fundamental capability for long-range perception systems, with broad value in remote sensing monitoring, maritime and aerial early warning, and security reconnaissance \cite{zhao2022single, kou2023infrared, yuan2026sp}. Compared to other imaging techniques, infrared sensors capture thermal radiation and can operate under weak illumination, nighttime conditions, and high disturbance. 
This passive imaging mechanism renders IRSTD uniquely effective in scenarios where it is difficult to observe targets through conventional visual cues. 
Leveraging these advantages, IRSTD offers significant practical value and is one of the most prominent and challenging tasks in low-level vision and remote sensing.

The difficulty of IRSTD stems from the inherent imaging characteristics of infrared small targets. 
Generally speaking, targets with fewer than 9$\times$9 pixels or occupying less than 0.15\% of the total image are considered small targets \cite{zhang2019infrared,zhang2021infrared}.
In long-range observations, a target often occupies only a few pixels, lacks stable shape and texture, and appears as an extremely sparse response embedded in a large background region \cite{zhao2022single,kou2023infrared}. Such targets provide insufficient semantic evidence for conventional object detection pipelines. 
Meanwhile, infrared backgrounds are often highly nonuniform: cloud edges, sensor noise, and other locally bright clutter may exhibit target-like contrast, leading to false alarms \cite{wang2019miss}.
Therefore, IRSTD relies more on reliable background suppression, weak target enhancement, and the precise distinction between sparse target responses and residual clutter signals \cite{chen2013local,gao2013infrared,dai2021attentional}. 

\begin{figure}[!t]
  \centering
   \includegraphics[width=0.95\linewidth]{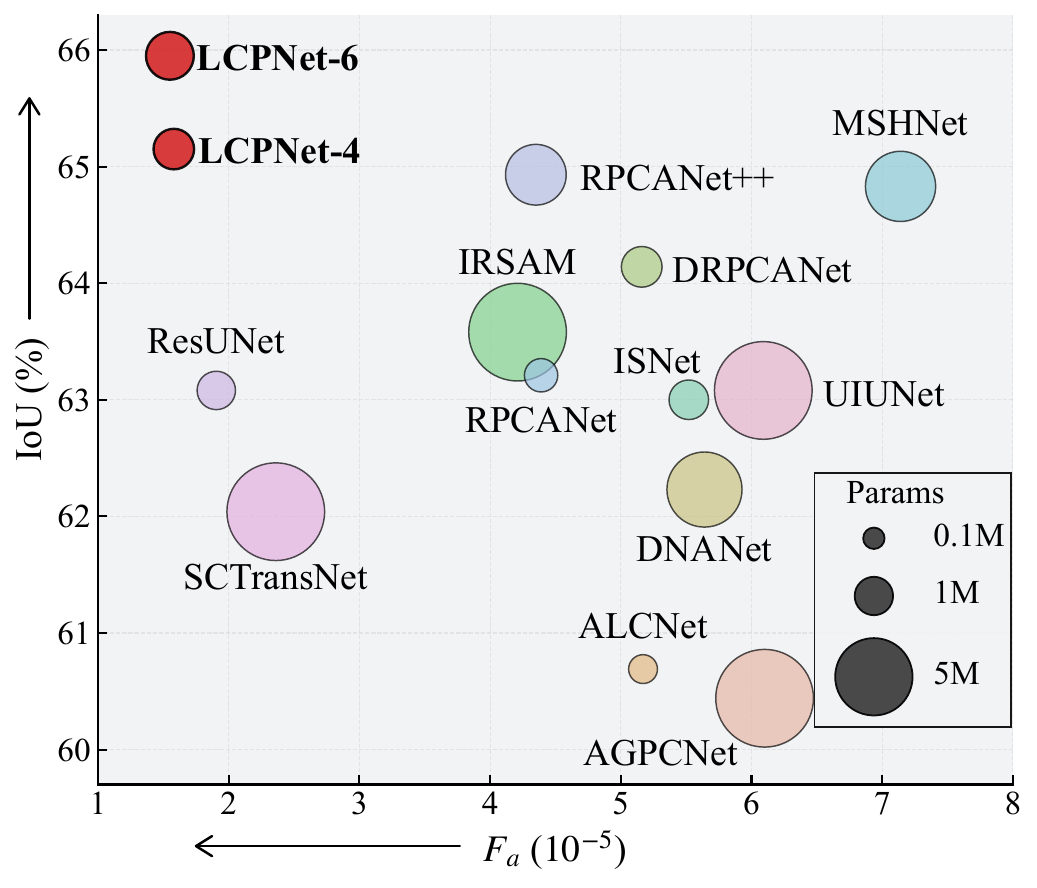}
  \caption{Comparison of $\text{F}_a(10^{-5})$-IoU scatter plots for infrared small target detection algorithms on IRSTD-1k \cite{zhang2022isnet}. Circle size indicates the parameter number. Points closer to top-left indicate better performance.}
  \label{fig:fa_iou_scatter}
\end{figure}

Recent deep learning methods have substantially advanced IRSTD by learning more powerful multiscale and context-aware representations from annotated infrared images. CNN-based segmentation models, U-shaped architectures, attention mechanisms, and dense nested networks have improved target localization and background discrimination under complex scenes \cite{wang2019miss,dai2021asymmetric,dai2021attentional,li2022dense,wu2022uiu,zhang2023attention,liu2024infrared}. These methods demonstrate that data-driven feature learning can compensate for the weak appearance of infrared small targets. 
However, most traditional deep learning models remain confined to an end-to-end learning paradigm that directly maps infrared images to target masks.
Under this paradigm, the learned behavior is mainly shaped by network architecture, loss supervision, and training data, rather than by an explicit physical model of infrared image formation. 
Consequently, it neglects the realistic physical modeling of background-target-noise decomposition.
This limitation causes models to generate excessive and cumulative false positives in complex scenarios.

\begin{figure}[t]
    \centering
    \includegraphics[width=\linewidth]{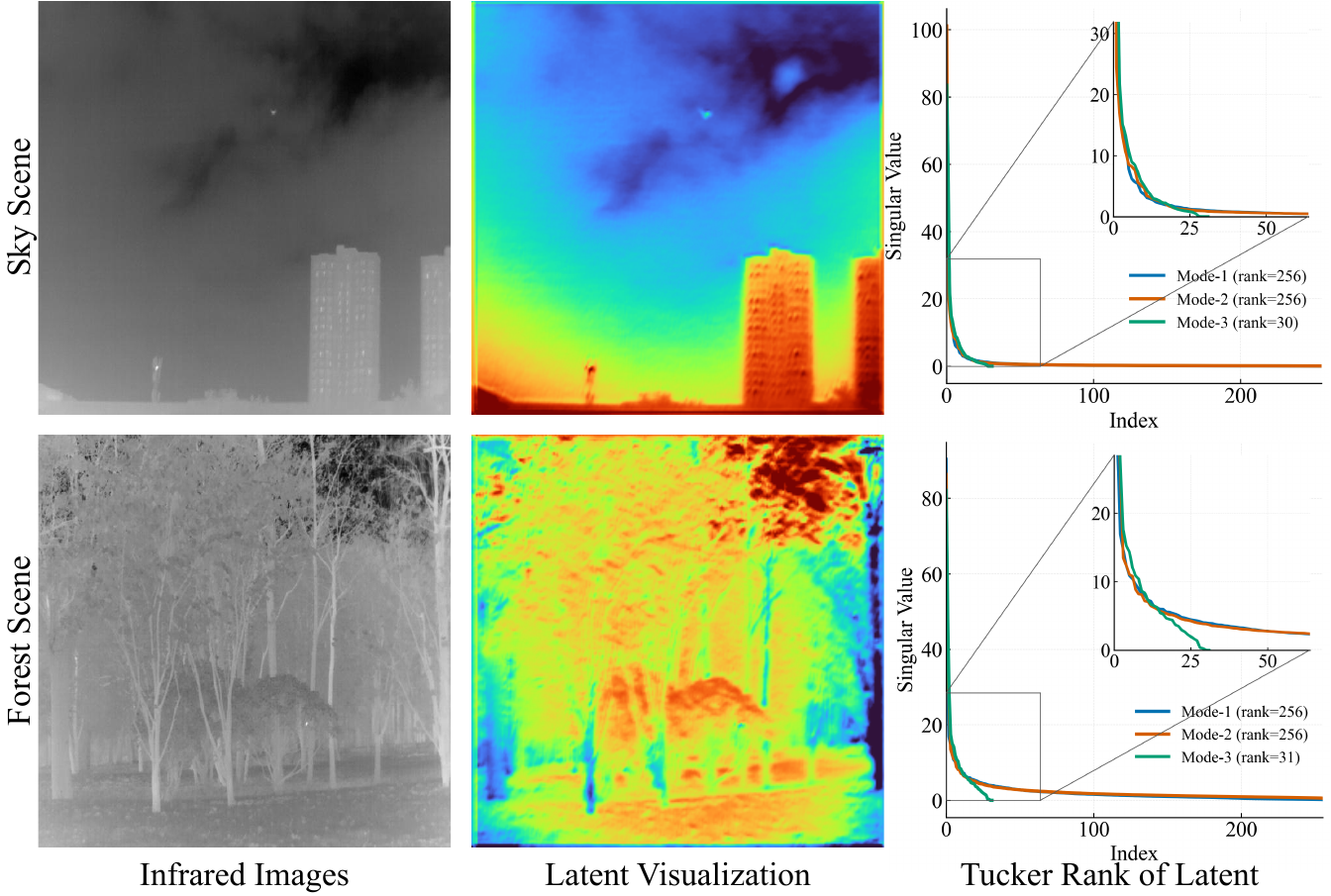}
    \caption{Visualization of the low-rank property in latent domain. From top to bottom are the original infrared image, a visualized channel from the latent, and Tucker rank analysis of the latent. It can be observed that, whether in simple scene (sky \textit{XDU82} in NUDT-SIRST \cite{li2022dense}) or complex scene (forest \textit{XDU41}), the latent exhibits distinct low-rank property in both visualization and Tucker rank analysis. For more scenes and rank analyses, please refer to Appendix~\ref{app:more_results}.}
    \label{fig:low-rank}
\end{figure}

Physical modeling for IRSTD motivates a review of the model-driven methods. Infrared small target image naturally admits a low-rank and sparse decomposition formulation, where the observation can be regarded as the superposition of a structured background, sparse target, and noise component, i.e., $\mathbf{X}=\mathbf{B}+\mathbf{T}+\mathbf{N}$ as shown in Fig.~\ref{fig:low-rank}. 
Classical RPCA and infrared patch-image models provide an interpretable foundation for separating sparse targets from low-rank backgrounds \cite{candes2011robust,gao2013infrared,zhang2018infrared,dai2017reweighted}. However, traditional optimization-based methods rely on fixed handcrafted priors, require iterative matrix computations, and are sensitive to regularization parameters and background statistics. Deep unfolding offers a principled compromise by unrolling iterative optimization into a finite number of learnable stages, thereby combining interpretable algorithmic structure with data-driven representation learning \cite{monga2021algorithm,wu2024rpcanet,wu2025rpcanet++}. This paradigm is particularly attractive for IRSTD because each stage can be associated with a target-background-noise separation process while still adapting the solver to real infrared data.

\begin{figure}[t]
    \centering
    \includegraphics[width=\linewidth]{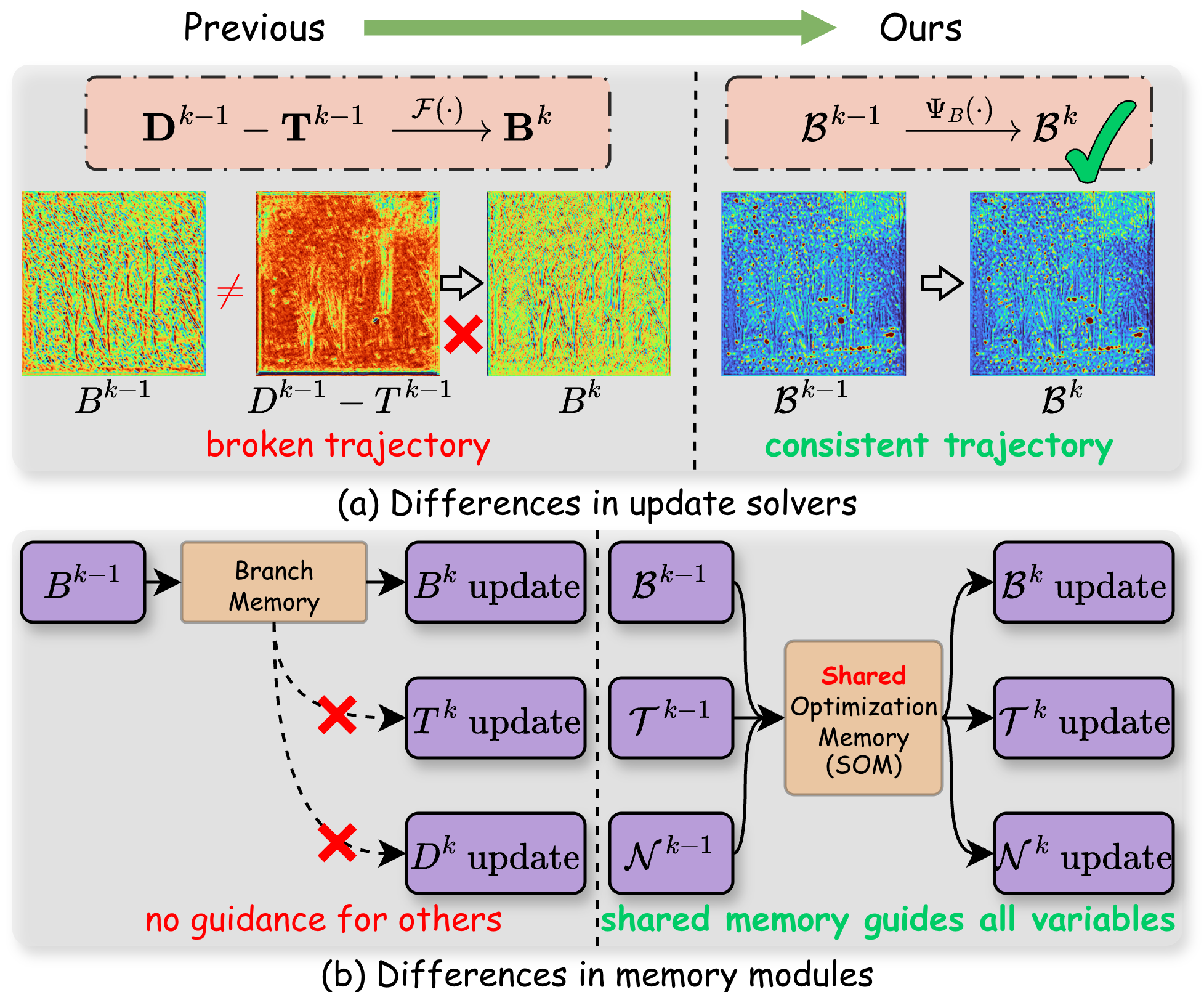}
    \caption{Illustration of design differences between methods. (a) Upper: formula-based modeling. Lower: feature visualization. In previous studies (left), $B^{k-1}$ and $B^{k}$ are similar, but the update using $D^{k-1}-T^{k-1}$ violates feature consistency. In our LCP solver (right), the current state is directly determined by previous state, thereby ensuring feature consistency. (b). Previous methods (left) only used branch-level memory when updating B. Our SOM (right) features a shared optimization memory state for all variables, which directly participates in all updates, ensuring the historical memory of each one.}
    \label{fig:lcp_solver_memory_compare}
\end{figure}

Despite this progress, existing deep unfolding methods still face three key limitations. 
First, to preserve the image-domain physical constraint, most unfolding frameworks transmit image-domain variables between adjacent stages \cite{wu2024rpcanet,wu2025rpcanet++}. This design requires each stage to project its intermediate representation back to low-dimensional images. Such repeated compression preserves explicit physical interface, but it would weaken the high-frequency and low-contrast responses that are critical for small target detection. 
Second, existing learned proximal solvers often obtain variables through indirect residual reconstruction, rather than directly evolving the previous state \cite{monga2021algorithm,wu2024rpcanet,wu2025rpcanet++} as shown in Fig.~\ref{fig:lcp_solver_memory_compare}(a). This weakens the continuity of background trajectory and may cause target leakage. Moreover, batch normalization is not well suited to IRSTD, since its statistics are dominated by abundant background pixels and fluctuate under small training batches, which can dilute weak target responses and destabilize repeated unfolding updates \cite{ioffe2015batch,wu2018group,cisse2017parseval,miyato2018spectral}.
Third, existing cross-stage memory mechanisms are typically attached to a single decomposition variable, so historical information is reused only within a local update \cite{shi2015convolutional,wu2025rpcanet++}. Such branch-specific memory does not represent a shared optimization state of the unfolding process as in Fig.~\ref{fig:lcp_solver_memory_compare}(b)(left), and therefore cannot provide common historical guidance for all coupled decomposition variables.

To address these issues, we propose Latent Consistent Proximal Unfolding Network (LCPNet) for infrared small target detection. 
After verifying that the low-rank physical prior remains valid in latent representation, we lift the execution domain of unfolding from image to latent, preserving the decomposition constraint while avoiding repeated projection of intermediate states. 
Then, LCPNet derives a Latent Consistent Proximal (LCP) solver for direct proximal updates, where each variable is evolved from its previous state rather than reconstructed from an indirect residual. Its updater further uses group-wise normalization and spectral gain control to better fit the small-batch and background-dominated characteristics of IRSTD. 
Finally, LCPNet replaces branch-specific memory with Shared Optimization Memory (SOM), which is accessible to all decomposition variables and provides unified guidance for the coupled latent optimization across stages.

The main contributions of this work are summarized as follows:
\begin{itemize}
    \item We verify that the low-rank decomposition prior remains valid in latent representations. Based on this observation, we formulate latent-domain unfolding to preserve the physical constraint while avoiding repeated variable compression.
    \item We derive a LCP solver for consistent proximal-state updates, allowing each latent variable to evolve from its previous state instead of indirect residual reconstruction. A group-wise and gain-controlled updater is further adopted to suppress background induced false alarms.
    \item We propose SOM as a system-level historical state for the unfolding process. SOM is shared by all decomposition variables and provides unified guidance for latent updates across stages.
    \item We propose LCPNet for infrared small target detection and extensive experiments on multiple IRSTD benchmarks demonstrate its superior detection accuracy, robust scene adaptability, and effective false-alarm suppression under diverse infrared scenes as shown in Fig.~\ref{fig:fa_iou_scatter}.
\end{itemize}










\begin{figure*}[!t]
  \centering
   \includegraphics[width=0.99\linewidth]{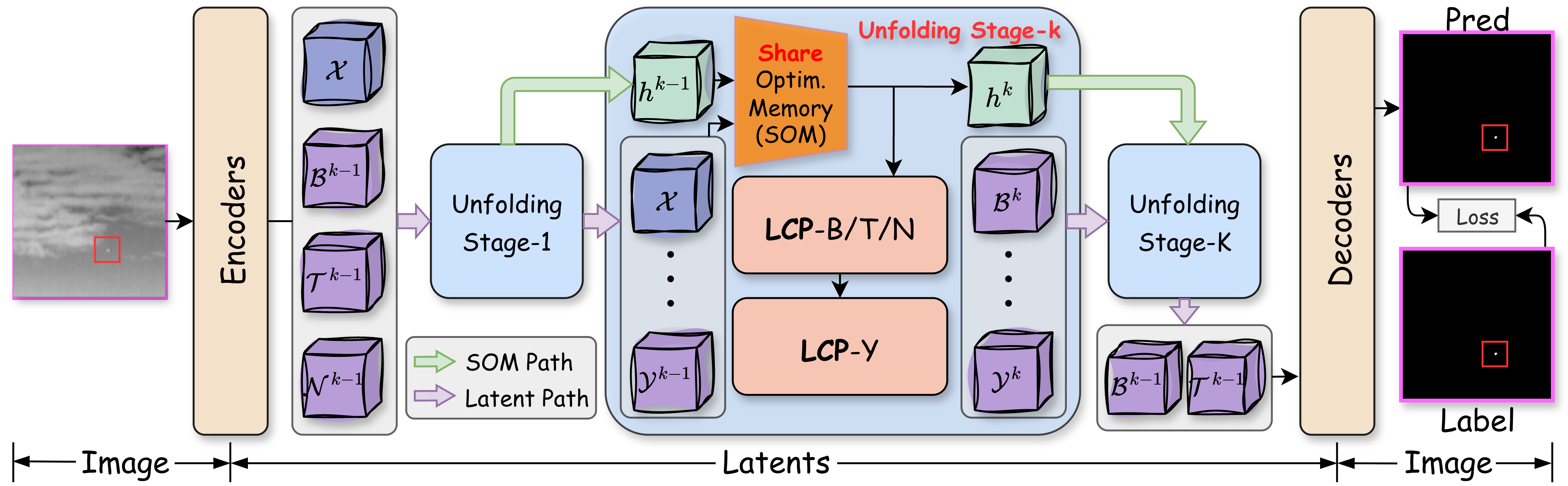}
  \caption{Overall architecture of LCPNet. An infrared image is first lifted into the latent domain, where latent-space decomposition is performed through $K$ unfolding stages, and the final target representation is decoded to image domain for detection.}
  \label{fig:arch}
\end{figure*}

\section{Related Work}

\subsection{Infrared Small Target Detection}

Existing IRSTD methods can be broadly grouped into Human Visual System (HVS)-based methods, optimization-based methods, and deep learning-based methods \cite{zhao2022single,kou2023infrared,peng2019infrared}. HVS-based methods are among the earliest solutions. They assume that small targets exhibit local intensity discontinuity relative to their surrounding background and therefore enhance target-like responses through handcrafted filters or local contrast measures. Representative methods include Top-hat filtering \cite{tom1993morphology}, max-mean and max-median filtering \cite{deshpande1999max}, local contrast measure (LCM) \cite{chen2013local}, and multiscale patch-based contrast measure (MPCM) \cite{wei2016multiscale}. These methods are simple, and computationally attractive, but their handcrafted saliency assumptions are often fragile in complex scenes.

Optimization-based methods regard infrared backgrounds as containing strong structural redundancy and can be represented by low-rank priors, while small targets are sparse outliers embedded in the observation. Based on this idea, the infrared patch-image (IPI) model formulates small target detection as a low-rank and sparse decomposition problem \cite{gao2013infrared}. NRAM further introduces non-convex rank approximation with structured sparsity to better characterize target components \cite{zhang2018infrared}, while PSTNN exploits patch-tensor modeling with nonlocal priors for single-frame detection \cite{dai2017reweighted}. Compared with HVS-based methods, optimization-based methods provide clearer physical interpretability. However, they usually require iterative solvers, manually designed priors, and carefully tuned parameters, which limits their efficiency.

Deep learning-based methods have recently become the dominant paradigm for IRSTD. Most of these methods are non-unfolding networks, which directly learn a segmentation mapping from the infrared input to the target mask in an end-to-end manner. Early data-driven studies such as MDvsFA treat IRSTD as small object segmentation and consider the tradeoff between missed detections and false alarms \cite{wang2019miss}. Subsequent works improve representation learning by designing IRSTD-specific networks, including asymmetric contextual modulation \cite{dai2021asymmetric}, attentional local contrast learning \cite{dai2021attentional}, dense nested attention \cite{li2022dense}, UNet-in-UNet structures \cite{wu2022uiu}, attention-guided pyramid context aggregation \cite{zhang2023attention}, and shape-aware supervision \cite{zhang2022isnet}. More recent methods further explore scale and location sensitivity \cite{liu2024infrared}, SAM adaptation for infrared small targets \cite{zhang2024irsam}, learnable saliency kernels \cite{wu2024saliency}, pinwheel-shaped convolution and scale-aware loss design \cite{yang2025pinwheel}, and low-level feature modeling \cite{li2025ilnet}. These networks have greatly improved detection accuracy, but their inference process is usually a black-box input-to-mask mapping.

Another emerging branch is deep unfolding-based IRSTD, which converts iterations of a traditional optimization algorithm into trainable stages \cite{monga2021algorithm}. This paradigm preserves model interpretability while replacing expensive analytical subsolvers with learnable neural modules. RPCANet and RPCANet++ instantiate this idea by unfolding RPCA-like decomposition for infrared small target detection and enhancing it with learnable stage-wise modules \cite{wu2024rpcanet,wu2025rpcanet++}. With its ability to integrate physical priors and data-adaptive learning, deep unfolding provides a promising foundation for more reliable and interpretable IRSTD models.

\subsection{Deep Unfolding in Vision Models}

Deep unfolding transforms an iterative optimization algorithm into a trainable network by mapping each iteration to a network stage and replacing fixed algorithmic components with learnable parameters or neural operators \cite{monga2021algorithm,sohldickstein2015deep,song2019generative,ho2020denoising,nichol2021improved,song2020denoising}. It bridges model-driven optimization and modern deep vision by preserving algorithmic structure and reducing iterative inference cost.

The idea first became influential in sparse coding and compressive sensing \cite{gregor2010learning,zhang2023optimization}. ISTA-Net unfolds ISTA into a structured network with learnable transforms, achieving interpretable and efficient reconstruction \cite{zhang2018ista}, while LDAMP and AMP-Net unfold denoising-based approximate message passing for compressive image recovery \cite{metzler2017learned,zhang2021ampnet}. ADMM-Net similarly unfolds ADMM for MRI, replacing shrinkage functions with trainable counterparts \cite{yang2016deep}.
Beyond this, unfolding has been widely adopted in image inverse problems \cite{aggarwal2018modl,hammernik2017learning}. These methods avoid learning the entire inverse mapping from scratch; instead, they constrain certain calculations using known physical laws, while neural modules learn the missing prior \cite{adler2018learned,sriram2020end}.
Low-rank and sparse decomposition is another natural domain for unfolding \cite{candes2011robust}. Learned RPCA extends this idea by unfolding scalable RPCA iterations and learning thresholds \cite{cai2021learned}. For IRSTD, RPCANet and RPCANet++ instantiate this principle and retain the background-target decomposition structure \cite{wu2024rpcanet,wu2025rpcanet++}.


\begin{figure*}[!t]
  \centering
   \includegraphics[width=0.99\linewidth]{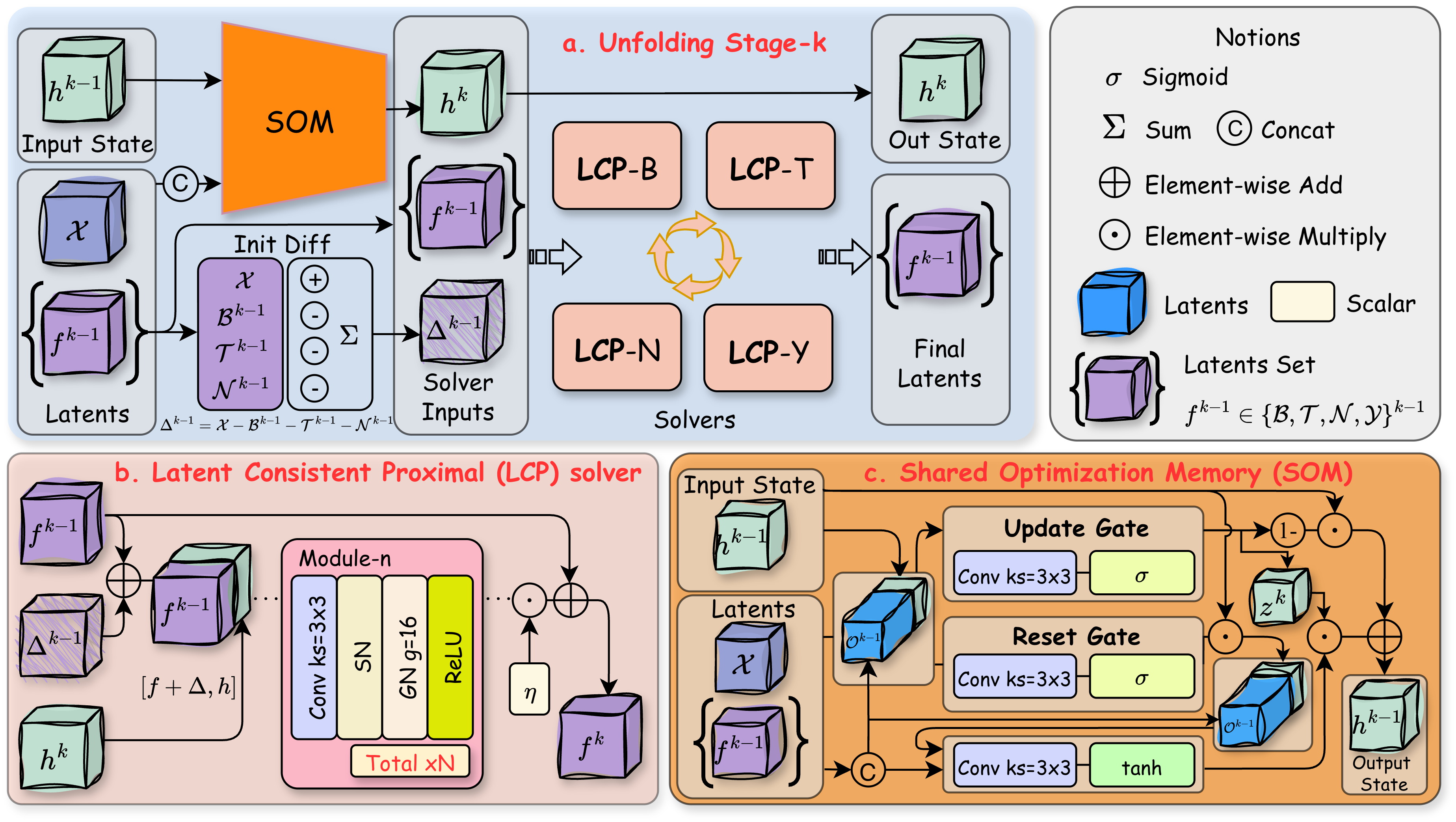}
  \caption{Details of our custom-designed modules. (a) Illustration of the $k$-th unfolding stage, $f^{k-1} \in \{ \mathcal{B}, \mathcal{T}, \mathcal{N}, \mathcal{Y} \}^{k-1}$ is a set of latent variables. (b) Illustration of LCP solver, each solver contains $N$ convolutional modules. $ks$ denotes convolutional kernel size, and $g$ denotes the group number in GN. (c) Illustration of SOM, where $O^{k-1}=Cat([\mathcal{X}, \{f^{k-1}\}])$.}
  \label{fig:stage_update_sosim}
\end{figure*}

\subsection{Recurrent and State-Space Optimization in Vision}

In unfolding vision networks, historical states help each stage reuse previous residuals along the optimization trajectory. This view is consistent with classical momentum and optimization \cite{nesterov2004introductory}, and is often implemented with recurrent modules such as LSTM, GRU, and ConvLSTM, which preserve temporal states through gated transitions \cite{hochreiter1997long,cho2014learning,shi2015convolutional}.

Recurrent state modeling has been adopted in several inverse-vision solvers. Recurrent inference machines learn iterative inverse-problem updates with a hidden state \cite{putzky2017recurrent} and MRI reuse cross-stage estimates through dense recurrent connections and recurrent variational networks \cite{hosseini2020dense,yiasemis2022recurrent,song2021memory}. These methods show that treating stage history as part of the solver state can improve reconstruction quality and optimization efficiency.
State-space models further broaden this idea for visual representation. S4 and Mamba propagate hidden states over long sequences with efficient structured \cite{gu2022s4,gu2024mamba}, while vision mamba adapt such state propagation to images through bidirectional visual scans \cite{zhu2024visionmamba,liu2024vmamba}. 
Motivated by these developments, LCPNet uses a shared optimization memory to represent the historical trajectory of the coupled latent decomposition variables, allowing system-level state to guide all primal-variable updates across unfolding stages.

\section{Methodology}

This section presents the methodology of LCPNet in a progressive manner. We first verified that physical priors remain valid in latents and reformulate the decomposition in latent domain. We then derive LCP solver for direct proximal-state updates, followed by the SOM that provides historical guidance for all coupled variables. Finally, we describe how these components are instantiated in the network architecture.

\subsection{Latent-Space Decomposition Formulation}

Classical low-rank and sparse decomposition methods formulate IRSTD as an image-domain separation problem. Given an infrared observation $\mathbf{X}$, the background, target, and noise components are estimated directly in the physical image space:
\begin{equation}
    \mathbf{X}=\mathbf{B}+\mathbf{T}+\mathbf{N}, \quad \text{s.t.} \quad \mathbf{X},\mathbf{B},\mathbf{T},\mathbf{N}\in\mathbb{R}^{H\times W}.
\end{equation}

This formulation is the basis of representative model-driven methods such as IPI, NRAM, and PSTNN \cite{gao2013infrared,zhang2018infrared,dai2017reweighted}, and it is also inherited by recent image-domain unfolding frameworks such as RPCANet and RPCANet++ \cite{wu2024rpcanet,wu2025rpcanet++}. Its main advantage is that the optimization variables remain directly interpretable in the physical image domain. However, when this formulation is unfolded into a deep network, each stage still exchanges single-channel physical variables and repeatedly projects the optimization trajectory back to $\mathbb{R}^{H\times W}$. This low-dimensional stage interface can attenuate high-frequency small-target clues during reconstruction and prevents background, target, and noise representations from evolving persistently in a high-dimensional feature space.

To address these limitations, we first verified the low-rank property of latent variable through visualization and singular value analysis, as shown in Fig. \ref{fig:low-rank}. The singular values plot of the feature tensor decay rapidly, and indicates that latent representation still satisfies the low-rank physical assumption.  This prior is applicable to both simple and complex scenarios, and Appendix~\ref{app:more_results} further demonstrates this prior from multiple perspectives.
More specifically, we use a set of encoders to map image-domain variables to latent variables:
\begin{equation}
    \begin{gathered}
    \mathcal{X}=E_X(\mathbf{X}), \quad \mathcal{B}=E_B(\mathbf{B}), \\
    \mathcal{T}=E_T(\mathbf{T}), \quad \mathcal{N}=E_N(\mathbf{N}).
    \end{gathered}
\end{equation}

The decomposition constraint is then imposed in the latent space:
\begin{equation}
    \mathcal{X}=\mathcal{B}+\mathcal{T}+\mathcal{N}, \quad \text{s.t.} \quad \mathcal{X},\mathcal{B},\mathcal{T},\mathcal{N}\in\mathbb{R}^{H\times W\times C}.
\end{equation}

Compared with the image-domain formulation, the latent manifold provides a high-dimensional stage interface, allowing small-target and background structures to be propagated through $C$-channel optimization variables. This design preserves weak high-frequency target evidence across stages and lets background, target, and noise representations evolve on a richer latent manifold. After the iterative decomposition, projectors map the target-related latent variables back to the image domain for final detection output.

Under this latent-space physical model, the general decomposition objective of IRSTD can be written as
\begin{equation}
    \mathcal{L}
    = \underbrace{\mathcal{R}_{\mathcal{B}}(\mathcal{B})+\mathcal{R}_{\mathcal{T}}(\mathcal{T})+\mathcal{R}_{\mathcal{N}}(\mathcal{N};\sigma)}_{\text{Constraint Items}} + \underbrace{\mathcal{F}(\mathcal{B},\mathcal{T},\mathcal{N})}_{\text{Fidelity Item}},
\end{equation}
where $\mathcal{R}_{\mathcal{B}}(\cdot)$, $\mathcal{R}_{\mathcal{T}}(\cdot)$, and $\mathcal{R}_{\mathcal{N}}(\cdot;\sigma)$ denote the constraint items, and these constraint functions are assumed to be unknown in the latent space. The fidelity item enforces consistency with the latent observation and is usually instantiated as $\mathcal{F}(\mathcal{B},\mathcal{T},\mathcal{N})=\frac{\mu}{2}\left\|\mathcal{X}-\mathcal{B}-\mathcal{T}-\mathcal{N}\right\|_F^2$.

Following the Alternating Direction Method of Multipliers (ADMM) \cite{boyd2011distributed}, we introduce the latent dual variable $\mathcal{Y}$ and write the augmented Lagrangian as:
\begin{equation}
    \begin{aligned}
        \mathcal{L}&(\mathcal{B},\mathcal{T},\mathcal{N},\mathcal{Y})
        =\mathcal{R}_{\mathcal{B}}(\mathcal{B})+\mathcal{R}_{\mathcal{T}}(\mathcal{T})+\mathcal{R}_{\mathcal{N}}(\mathcal{N};\sigma) \\
        &+\left\langle \mathcal{Y},\mathcal{X}-\mathcal{B}-\mathcal{T}-\mathcal{N}\right\rangle+\frac{\mu}{2}\left\|\mathcal{X}-\mathcal{B}-\mathcal{T}-\mathcal{N}\right\|_F^2.
    \end{aligned}
\label{eq:latent_augmented_lagrangian}
\end{equation}

ADMM decomposes this objective into three main subproblems for $\mathcal{B}$, $\mathcal{T}$, and $\mathcal{N}$. Since these subproblems share the same proximal form, we take the $\mathcal{B}$-subproblem as the representative case, with the detailed derivation provided in Appendix~\ref{app:admm_subproblem_derivation}:
\begin{equation}
    \begin{aligned}
        \mathcal{B}^{k}=&\arg\min_{\mathcal{B}}\ \mathcal{R}_{\mathcal{B}}(\mathcal{B}) \\
        &+\frac{\mu}{2}\left\|\mathcal{X}-\mathcal{B}-\mathcal{T}^{k-1}-\mathcal{N}^{k-1}+\frac{1}{\mu}\mathcal{Y}^{k-1}\right\|_F^2.
    \end{aligned}
\end{equation}

The next subsection details how LCPNet turns this implicit latent proximal step into a direct learnable solver.


\subsection{Latent Consistent Proximal (LCP) Solver}

We derive the LCP solver from the $\mathcal{B}$-subproblem. For compactness, define the problem as:
\begin{equation}
    \mathcal{V}_{B}^{k-1}=\mathcal{X}-\mathcal{T}^{k-1}-\mathcal{N}^{k-1}+\frac{1}{\mu_B}\mathcal{Y}^{k-1},
\label{eq:VB_definition}
\end{equation}
where $\mu_B$ denotes the $\mathcal{B}$-specific augmented-Lagrangian penalty coefficient.
Then the $\mathcal{B}$ update can be written in the proximal form
\begin{equation}
    \mathcal{B}^{k}=\arg\min_{\mathcal{B}}\ \mathcal{R}_{\mathcal{B}}(\mathcal{B})+\frac{\mu_B}{2}\left\|\mathcal{B}-\mathcal{V}_{B}^{k-1}\right\|_F^2.
\label{eq:B_proximal_subproblem}
\end{equation}

In conventional optimization-based IRSTD models, the regularizer $\mathcal{R}_{\mathcal{B}}(\cdot)$ is manually specified to encode a prescribed background prior, such as low-rank priors with nuclear norm $\| \cdot \|_*$ constraints. 
However, there is an inherent gap between manually specified priors and the true physical attributes, so no analytical constraint can fully characterize the desired component. We therefore regard $\mathcal{R}_{\mathcal{B}}(\cdot)$ as an unknown latent regularizer.
Specifically, assume that $\nabla\mathcal{R}_{\mathcal{B}}$ is $L_B$-Lipschitz continuous:
\begin{equation}
    \left\|\nabla\mathcal{R}_{\mathcal{B}}(\mathcal{U})-\nabla\mathcal{R}_{\mathcal{B}}(\mathcal{V})\right\|_F\leq L_B\left\|\mathcal{U}-\mathcal{V}\right\|_F.
\label{eq:RB_lipschitz_assumption}
\end{equation}

Under the Lipschitz condition in Eq.~\ref{eq:RB_lipschitz_assumption}, the smooth Taylor upper bound \cite{nesterov2004introductory} gives:
\begin{equation}
    \begin{aligned}
        \mathcal{R}_{\mathcal{B}}(\mathcal{B})\leq &\ \mathcal{R}_{\mathcal{B}}(\mathcal{B}^{k-1})+\left\langle\nabla\mathcal{R}_{\mathcal{B}}(\mathcal{B}^{k-1}),\mathcal{B}-\mathcal{B}^{k-1}\right\rangle \\
        &+\frac{L_B}{2}\left\|\mathcal{B}-\mathcal{B}^{k-1}\right\|_F^2,\quad \forall \mathcal{B}\in\mathbb{R}^{H\times W\times C}.
    \end{aligned}
\label{eq:RB_taylor_majorization}
\end{equation}

Substituting the majorization in Eq.~\ref{eq:RB_taylor_majorization} into the proximal subproblem in Eq.~\ref{eq:B_proximal_subproblem}, and removing constants independent of $\mathcal{B}$, gives the surrogate objective:
\begin{equation}
    \begin{aligned}
        \hat{\mathcal{J}}_{\mathcal{B}}(\mathcal{B})=&\ \frac{L_B}{2}\left\|\mathcal{B}-\left(\mathcal{B}^{k-1}-\frac{1}{L_B}\nabla\mathcal{R}_{\mathcal{B}}(\mathcal{B}^{k-1})\right)\right\|_F^2 \\
        &+\frac{\mu_B}{2}\left\|\mathcal{B}-\mathcal{V}_{B}^{k-1}\right\|_F^2.
    \end{aligned}
\label{eq:B_surrogate_objective}
\end{equation}

The surrogate objective in Eq.~\ref{eq:B_surrogate_objective} is the sum of two quadratic terms. Its minimizer admits the following closed-form update, with the detailed derivation provided in Appendix~\ref{app:admm_surrogate_derivation}:
\begin{equation}
    \mathcal{B}^{k}
    =\frac{L_B}{L_B+\mu_B}\mathcal{B}^{k-1}
    +\frac{\mu_B}{L_B+\mu_B}\mathcal{V}_{B}^{k-1}
    -\frac{\nabla \mathcal{R}_{\mathcal{B}}(\mathcal{B}^{k-1})}{L_B+\mu_B}.
\label{eq:B_closed_form}
\end{equation}

The closed-form solution in Eq.~\ref{eq:B_closed_form} is strictly equivalent to a gradient-like descent step:
\begin{equation}
    \mathcal{B}^{k}=\mathcal{B}^{k-1}-\frac{1}{L_B+\mu_B}\Big(\nabla\mathcal{R}_{\mathcal{B}}(\mathcal{B}^{k-1})+\mu_B(\mathcal{B}^{k-1}-\mathcal{V}_{B}^{k-1})\Big).
\label{eq:B_gradient_step_equivalent}
\end{equation}

Equivalently, this update can be written as:
\begin{equation}
    \mathcal{B}^{k}=\mathcal{B}^{k-1}-\eta_B\mathcal{G}_{B}^{k-1},
\label{eq:B_direction_update}
\end{equation}
where $\eta_B=1/(L_B+\mu_B)$. 
This formulation explicitly anchors the update on the previous state $\mathcal{B}^{k-1}$, providing a more direct reference for estimating the next background state $\mathcal{B}^{k}$. By substituting Eq.~\ref{eq:VB_definition} into Eq.~\ref{eq:B_gradient_step_equivalent} and defining the current latent residual as $\mathcal{P}^{k-1}=\mathcal{X}-\mathcal{B}^{k-1}-\mathcal{T}^{k-1}-\mathcal{N}^{k-1}$, the ideal update direction is directly obtained as:
\begin{equation}
    \mathcal{G}_{B}^{k-1}=\nabla\mathcal{R}_{\mathcal{B}}(\mathcal{B}^{k-1})-\mu_B\mathcal{P}^{k-1}-\mathcal{Y}^{k-1}.
\label{eq:B_direction_residual_form}
\end{equation}

However, directly computing $\mathcal{G}_{B}^{k-1}$ inside a network module is not practical. First, $\mathcal{R}_{\mathcal{B}}(\cdot)$ is an unknown latent regularizer, so $\nabla\mathcal{R}_{\mathcal{B}}(\mathcal{B}^{k-1})$ has no explicit analytical form. Second, explicitly concatenating $\mathcal{Y}^{k-1}$, $\mathcal{B}^{k-1}$, and $\mathcal{P}^{k-1}$ as solver inputs increases the input complexity of each unfolding stage. Third, the true update direction should not depend only on the current variables, it should also reflect the historical optimization trajectory across previous stages.
Therefore, we replace the ideal direction in Eq.~\ref{eq:B_direction_residual_form} with a learnable latent proximal surrogate:
\begin{equation}
    \mathcal{G}_{B}^{k-1} \approx \widetilde{\mathcal{G}}_{B}^{k-1}=\Psi_{B}(\mathcal{B}^{k-1}-\mathcal{P}^{k-1},h^{k}).
\label{eq:B_surrogate_direction}
\end{equation}

Instead of directly concatenating all solver variables, $\mathcal{B}^{k-1}-\mathcal{P}^{k-1}$ serves as a compact variable-centered residual that fuses the current background estimate and the decomposition-consistency violation into a single surrogate cue.
Based on this approximation, we obtain a more direct Markov-like LCP solver that incorporates historical trajectories:
\begin{equation}
    \mathcal{B}^{k}=\mathcal{B}^{k-1} - \eta_B \Psi_{B}(\mathcal{B}^{k-1}-\mathcal{P}^{k-1}, h^{k}),
\label{eq:lcp_solver}
\end{equation}
$h^{k}$ complements the current variables with global memory from previous stages. The construction of this shared state and its role in encoding historical optimization trajectories are detailed in Section \ref{subsec:som}.

\subsection{Shared Optimization Memory (SOM)}
\label{subsec:som}

Existing recurrent unfolding methods usually place the memory mechanism inside a specific branch of the unfolding network. For example, RPCANet++ introduces a memory-augmented module to alleviate inter-stage transmission loss in background approximation module, mainly aiming to preserve background features across stages \cite{wu2025rpcanet++}. The stored memory mainly serves the background branch, and thus cannot directly modulate the target and noise updates. The target branch can only benefit from this memory indirectly through the background representation. 
Moreover, the variables $\mathcal{B}$, $\mathcal{T}$, $\mathcal{N}$, $\mathcal{Y}$, and $\mathcal{X}$ are not encoded as a coupled decomposition system, so the historical optimization trajectory is only partially represented.

To address these limitations, LCPNet constructs a SOM for the whole decomposition process. At stage $k$, we form a joint observation by concatenating the current latent variables:
\begin{equation}
    o_{k-1}=\Big[\mathcal{B}^{k-1},\mathcal{T}^{k-1},\mathcal{N}^{k-1},\mathcal{Y}^{k-1},\mathcal{X}\Big].
\label{eq:joint_optimization_observation}
\end{equation}

An optimization state cell then maps this joint observation into a shared state $h_k$ through a gated recurrent mechanism \cite{cho2014learning}:
\begin{equation}
    \begin{gathered}
        z_k=\sigma(W_z*[o_{k-1},h_{k-1}]+b_z),\\
        r_k=\sigma(W_r*[o_{k-1},h_{k-1}]+b_r),\\
        \tilde{h}_k=\tanh(W_q*[o_{k-1},r_k\odot h_{k-1}]+b_q),\\
        h_k=(1-z_k)\odot h_{k-1}+z_k\odot\tilde{h}_k.
    \end{gathered}
\label{eq:optimization_state_cell}
\end{equation}

As shown in Fig \ref{fig:stage_update_sosim}(c), $*$ denotes convolution, $z_k$ is the update gate, $r_k$ is the reset gate, and $\odot$ denotes element-wise multiplication. Unlike branch-specific memory, $h_k$ is shared by the $\mathcal{B}$, $\mathcal{T}$, and $\mathcal{N}$ update branches and is directly fed into their proximal surrogate updaters. Therefore, the same system-level state can modulate all primal updates, making $h_k$ a global control variable for the unfolding solver.

The adaptive gates aggregate historical decomposition observations into a high-order effective memory of the unfolding trajectory. Recursively expanding Eq.~\ref{eq:optimization_state_cell} gives
\begin{equation}
    h_k=\left(\prod_{i=1}^{k}(1-z_i)\right)\odot h_0+\sum_{t=1}^{k}\left[z_t\odot\tilde{h}_t\odot\prod_{i=t+1}^{k}(1-z_i)\right].
\label{eq:optimization_state_expansion}
\end{equation}

Eq.~\ref{eq:optimization_state_expansion} shows that $h_k$ is a gated accumulation of all previous candidate states, where historical information is adaptively retained across stages. The detailed proof is provided in Appendix~\ref{app:sim_memory_expansion}, and this expansion indicates that SOM forms a shared high-order memory for the whole latent system rather than isolated branch-wise states.

\subsection{Network Architecture}

The overall architecture of LCPNet is shown in Fig.~\ref{fig:arch}. Given an infrared image, LCPNet first lifts the input into latent domain and then performs $K$ stages of latent decomposition unfolding. As illustrated in Fig.~\ref{fig:stage_update_sosim}(a), each stage updates the background, target, noise, and dual variables under the latent decomposition constraint, while SOM provides historical guidance for the stage-wise proximal surrogate updates. The final target representation is decoded back to the image domain to generate the detection map.

The update module in each stage is shown in Fig.~\ref{fig:stage_update_sosim}(b). Following the LCP solver derived in Eq. \ref{eq:B_surrogate_direction}, the module estimates the learned proximal direction from the variable residual and shared optimization state. The same design is used for different primal variable branches. 

For the updater in LCP solver, a standard batch normalization design is sensitive to mini-batch statistics, which is not ideal for IRSTD because infrared backgrounds vary across scenes and small targets occupy only a few pixels \cite{ioffe2015batch}. We therefore use Group Normalization (GN) to normalize features within each sample, reducing the dependence on batch-level statistics while preserving scene-specific feature variations \cite{wu2018group}. 
Furthermore, Spectral Normalization (SN) is adopted to constrain the gain of the convolutional updater, preventing background fluctuations from being repeatedly amplified and improving the discrimination of weak target responses:
\begin{equation}
    \operatorname{SN}(\mathbf{W}) = \frac{\mathbf{W}}{\sigma_{\max}(\mathbf{W})},
\label{eq:spectral_normalization}
\end{equation}
where $\sigma_{\max}(\mathbf{W})$ denotes the largest singular value of the convolutional weight $\mathbf{W}$, and the normalized operator is used in the recurrent unfolding updater \cite{miyato2018spectral}.
As shown in Fig.~\ref{fig:stage_update_sosim}(b), the final update module combines this gain-controlled convolution with sample-wise normalization and nonlinear activation, providing a smoother approximation to the learned proximal direction $\widetilde{\mathcal{G}}_{B}^{k-1}$.

\begin{figure*}[!t]
  \centering
   \includegraphics[width=0.9\linewidth]{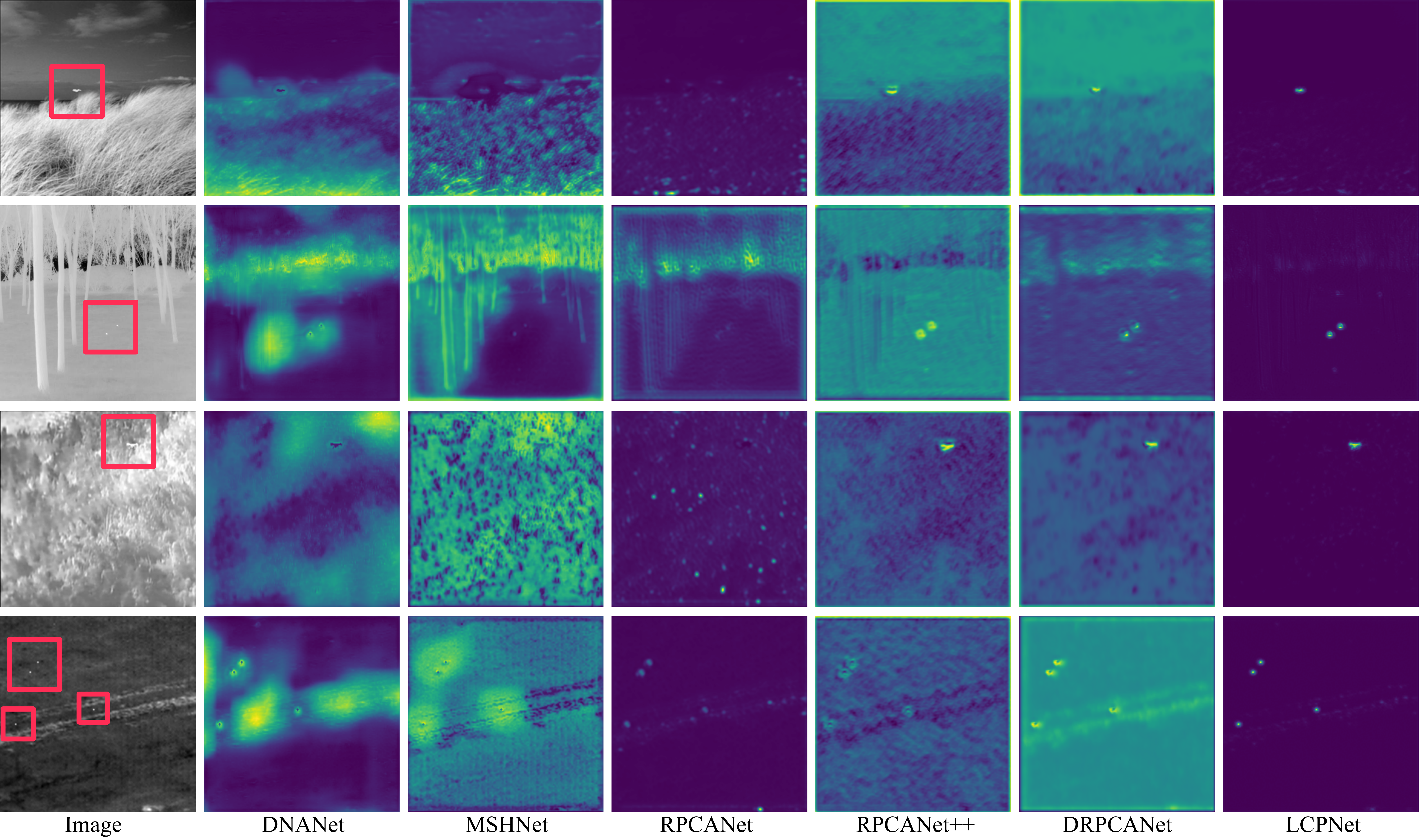}
  \caption{Visualization of the last layer feature maps produced by different methods. For each method, the feature map is projected by PCA, and the most informative principal component is selected for visualization. Rows 1, 3, 4 are \textit{001322}, \textit{000562}, \textit{000377} in NUDT-SIRST \cite{li2022dense}. Row 2 is \textit{XDU902} in IRSTD-1k \cite{zhang2022isnet}.}
  \label{fig:heatmap_1}
\end{figure*}

\begin{table*}[!ht]
    \renewcommand\arraystretch{1.3}
    \tabcolsep=0.3mm
    \begin{center}
    \caption{Comparison with state-of-the-art IRSTD methods in terms of IoU(\%), $\text{F}_1$(\%), $\text{P}_d$(\%) and $\text{F}_a(10^{-5})$ on four benchmarks. The best and second-best results under each metric are highlighted in \textbf{bold} and \underline{underlined}, respectively. Inference speed is measured on an NVIDIA GeForce RTX 3090 GPU and an AMD EPYC 7502 CPU.}
    \label{table:sota}
    \resizebox{\linewidth}{!}{
    
        \begin{tabular}{l|cccc|cccc|cccc|cccc|ccc}
        \TableOuterRule
        \multirow{2}{*}{Model} & \multicolumn{4}{c|}{NUDT-SIRST} & \multicolumn{4}{c|}{IRSTD-1K} & \multicolumn{4}{c|}{SIRST} & \multicolumn{4}{c|}{SIRST-Aug} & Params & Flops & Time(ms) \\
        & IoU$\uparrow$ & $\text{F}_1$$\uparrow$ & $\text{P}_d$$\uparrow$ & $\text{F}_a$$\downarrow$ & IoU$\uparrow$ & $\text{F}_1$$\uparrow$ & $\text{P}_d$$\uparrow$ & $\text{F}_a$$\downarrow$ & IoU$\uparrow$ & $\text{F}_1$$\uparrow$ & $\text{P}_d$$\uparrow$ & $\text{F}_a$$\downarrow$ & IoU$\uparrow$ & $\text{F}_1$$\uparrow$ & $\text{P}_d$$\uparrow$ & $\text{F}_a$$\downarrow$ & (M) & (G) & (CPU/GPU) \\

        \TableInnerRule
        \multicolumn{20}{c}{\textit{\textbf{HVS-Based Methods}}} \\
        \TableInnerRule
        \SotaAltRow Tophat \cite{tom1993morphology} & 22.25 & 36.41 & 91.21 & 75.73 & 6.64 & 12.46 & 81.09 & 14.24 & 26.93 & 42.44 & 88.99 & 18.27 & 16.37 & 28.14 & 81.15 & 160.3 & - & - & 0.08/- \\
        Max-Median \cite{deshpande1999max} & 20.49 & 34.01 & 84.97 & 31.27 & 20.76 & 34.38 & 85.56 & 21.21  & 32.56 & 49.12 & 94.46 & 3.38 & 16.88 & 28.89 & 79.36 & 8.25 & - & - & 7836/- \\
        \SotaAltRow MPCM \cite{wei2016multiscale} & 9.26 & 16.95 & 70.58 & 32.72 & 15.70 & 27.13 & 62.54 & 12.47 & 22.30 & 36.47 & 90.83 & 5.56 & 19.76 & 33.00 & 93.39 & 3.14 & - & - & 69.78/-  \\ 
        HBMLCM \cite{peng2019infrared} & 7.60 & 14.13 & 60.74 & 17.84 & 12.60 & 22.38 & 57.73 & 10.08 & 11.95 & 21.36 & 79.81 & 41.32 & 2.37 & 46.37 & 50.61 & 146.1 & - & - & 6107/- \\
        
        \TableInnerRule
        \multicolumn{20}{c}{\textit{\textbf{Optimization-Based Methods}}} \\
        \TableInnerRule

        \SotaAltRow IPI \cite{gao2013infrared} & 34.62 & 51.43 & 92.38 & 7.54 & 25.27 & 40.35 & 83.51 & 20.93 & 44.83 & 61.90 & 97.25 & 3.76 & 21.93 & 35.97 & 80.05 & 3.62 & - & - & 8004/- \\ 
        NRAM \cite{zhang2018infrared} & 11.44 & 20.52 & 65.71 & 2.35 & 17.97 & 30.46 & 59.79 & 5.71 & 26.39 & 41.75 & 88.07 & 1.60 & 10.41 & 18.86 & 76.06 & 3.26 & - & - & 2443/-  \\ 
        \SotaAltRow NOLC \cite{zhang2019infrared} & 18.60 & 31.37 & 70.68 & 3.96 & 10.90 & 19.65 & 64.26 & 1.83 & 20.35 & 33.82 & 83.48 & 1.36 & 8.68 & 15.97 & 59.00 & 6.69 & - & - & 102.3/- \\
        
        PSTNN \cite{dai2017reweighted} & 25.17 & 40.22 & 80.21 & 7.61 & 24.39 & 39.22 & 65.29 & 15.77 & 35.17 & 52.04 & 88.99 & 2.57 & 12.38 & 22.04 & 60.79 & 2.98 & - & - & 62.46/- \\ 
        
        \TableInnerRule
        \multicolumn{20}{c}{\textit{\textbf{Deep learning-Based Methods}}} \\
        \TableInnerRule

        \SotaAltRow ACM \cite{dai2021asymmetric} & 69.46 & 81.98 & 97.14 & 13.11 & 54.16 & 70.25 & 83.84 & 7.32 & 66.77 & 80.07 & 98.16 & 11.03 & 67.79 & 80.80 & 95.87 & 32.52 & 0.51 & 0.42 & -/7.17 \\ 
        AGPCNet \cite{zhang2023attention} & 85.02 & 91.90 & 97.88 & 4.34 & 60.44 & 75.34 & 90.72 & 6.10 & 68.19 & 81.08 & \underline{99.10} & 12.09 & 72.07 & 83.77 & 98.62 & 29.39 & 12.4 & 43.0 & -/14.71 \\
        \SotaAltRow DNANet \cite{li2022dense} & 87.73 & 93.47 & 97.67 & 5.52 & 62.23 & 76.72 & \underline{92.44} & 5.64 & 72.62 & 84.28 & 99.08 & 10.26 & 70.56 & 82.74 & 96.56 & 36.49 & 4.69 & 13.9 & -/22.64 \\ 
        UIUNet \cite{wu2022uiu} & 88.94 & 94.15 & 95.23 & 1.53 & 63.08 & 77.36 & 92.10 & 6.09 & 72.15 & 83.82 & 98.16 & 7.90 & 70.76 & 82.88 & 96.70 & 34.30 & 50.5 & 54.2 & -/27.50  \\ 
        \SotaAltRow MSHNet \cite{liu2024infrared} & 75.96 & 86.34 & 95.23 & 8.28 & 64.83 & 78.67 & \textbf{92.78} & 7.14 & 64.64 & 78.52 & 98.16 & 15.66 & 71.93 & 83.67 & \textbf{99.03} & 32.06 & 4.06 & 6.01 & -/13.52 \\     
        SCTransNet \cite{yuan2024sctransnet} & 89.49 & 94.45 & 97.67 & 1.02 & 62.04 & 76.56 & 73.47 & 2.36 & 71.15 & 83.14 & 97.24 & 0.89  & 72.05 & 83.20 & 97.76 & 5.20 & 11.3 & 10.0 & -/31.42 \\

        \SotaAltRow ALCNet \cite{dai2021attentional} & 80.75 & 89.35 & 97.56 & 9.56 & 60.69 & 75.53 & 85.57 & 5.17 & 70.31 & 82.57 & 99.08 & 11.36 & 67.38 & 80.51 & 97.94 & 28.68 & 0.43 & 0.38 & -/9.51 \\ 
        ISNet \cite{zhang2022isnet} & 87.05 & 93.08 & 96.83 & 4.05 & 63.00 & 77.29 & 88.66 & 5.52 & 66.96 & 80.20 & 96.33 & 9.85 & 71.10 & 83.11 & 97.66 & 30.33 & 1.08 & 30.3 & -/28.30 \\
        
        \SotaAltRow IRSAM \cite{zhang2024irsam} & 87.85 & 93.53  & \textbf{99.05} & 3.21 & 63.58  &  77.74&  92.10 & 4.21 &  67.97 & 80.93  & 99.08 & 8.34  & 72.29 & 83.95 & \underline{99.02} & 25.94 & 10.5 & 37.6 & -/75.06 \\ 
        L2SKNet \cite{wu2024saliency} &  93.08 &  96.42 &  97.88& 1.35  & 57.07 & 72.67  & 90.03 & 37.90 & 62.54  &  76.96& 97.25  &  10.63& 71.89 & 83.65 & 98.62 & 31.56 & 1.07 & 5.92 & -/9.38 \\ 
        
        \TableInnerRule
        \multicolumn{20}{c}{\textit{\textbf{Deep Unfolding-Based Methods}}} \\
        \TableInnerRule

        \SotaAltRow RPCANet \cite{wu2024rpcanet} & 89.31 & 94.35 & 97.14 & 2.87 & 63.21 & 77.45 & 88.31 & 4.39 & 70.37 & 82.61 & 95.41 & 7.42 & 72.54 & 84.08 & 98.21 & 34.14 & 0.68 & 44.3 & -/15.40 \\ 
        
        RPCANet++ \cite{wu2025rpcanet++} & 94.39 & 97.12 & \underline{98.41} & 1.34 & 64.93 & 78.73 & 89.70 & 4.35 & \textbf{74.63} & \textbf{85.47} & \textbf{100.0} & 10.22 & 69.97 & 82.33 & 98.76 & 28.00 & 2.96 & 190 & -/43.61 \\ 
        
        

        \SotaOursRow LCPNet-4 & \underline{95.70} & \underline{97.80} & 97.57 & \textbf{0.03} & \underline{65.15} & \underline{78.89} & 86.60 & \underline{1.58} & 71.18 & 83.17 & 94.50 & \textbf{0.87} & \underline{72.75} & \underline{84.22} & 94.64 & \textbf{2.10} & 1.15 & 75.2 & -/23.54  \\
        \SotaOursRow LCPNet-6 & \textbf{95.77} & \textbf{97.84} & 97.35 & \underline{0.09} & \textbf{65.95} & \textbf{79.48} & 87.63 & \textbf{1.55} & \underline{72.95} & \underline{84.36} & 96.33 & \underline{0.89} & \textbf{73.68} & \textbf{84.85} & 96.15 & \underline{2.24} & 1.72 & 112 & -/31.38  \\


        \TableOuterRule
        \end{tabular}
    }

    \end{center}
\end{table*}

\section{Experiments}

This section evaluates the proposed method from several complementary perspectives. 
We first introduce the datasets, metrics, and implementation settings, and then compare the proposed method with SOTAs in terms of quantitative accuracy, visual detection quality, and computational efficiency. 
Then, ablation studies are conducted to examine the contribution of each key design.
Finally, we discuss the differences and limitations of our method compared to other approaches, as well as our outlook for the future.

\subsection{Implementation Details}
\subsubsection{Datasets and Metrics}
\label{subsec:datasets_metrics}

We evaluate LCPNet on four publicly available benchmarks, namely NUDT-SIRST \cite{li2022dense}, IRSTD-1K \cite{zhang2022isnet}, SIRST \cite{dai2021asymmetric}, and SIRST-Aug \cite{zhang2023attention}. 
NUDT-SIRST \cite{li2022dense} contains 1327 images with a resolution of $256\times256$, and is widely used to evaluate robustness under diverse backgrounds and complex clutter conditions. Following the protocol of \cite{li2022dense}, the training and test sets are split 1:1. 
IRSTD-1K \cite{zhang2022isnet} is a realistic infrared small target detection benchmark, comprising manually annotated real-world infrared images. It contains 800 training images and 201 test images.
SIRST \cite{dai2021asymmetric} is one of the earliest and most widely used single-frame benchmarks. It contains 341 training images and 86 test images in various sizes.
SIRST-Aug \cite{zhang2023attention} is an augmented version derived from SIRST \cite{dai2021asymmetric} through data augmentation. It contains 8525 training images and 545 test images and provides more sufficient training samples for deep models. 

For quantitative evaluation, we adopt five commonly used metrics, including mean Intersection over Union (IoU), $F_{1}$ score, Probability of detection ($\text{P}_d$), and $\text{F}_a$lse alarm rate ($\text{F}_a$).
They assess performance from the perspectives of overall segmentation quality, the ability to correctly detect real objects, and the risk of false alarms in cluttered backgrounds, respectively.
Additionally, we included area under the ROC curve (AUC) to summarize the detection performance at different decision thresholds based on the ROC curve.


\subsubsection{Experimental Settings}
\label{subsec:experimental_settings}

All implementations were developed based on PyTorch. The experiments were conducted on a workstation equipped with an NVIDIA GeForce RTX 3090 GPU and an AMD EPYC 7502 CPU.
For training, all input images were resized to a base size of $256\times256$, and the batch size was set to $8$. The model was trained for $800$ epochs on NUDT-SIRST, IRSTD-1K, and SIRST, and for $400$ epochs on SIRST-Aug. We used the Adam optimizer with an initial learning rate of $1\times10^{-4}$, and adopted the poly learning rate scheduler with a power of $0.9$. The network is supervised by SoftIoU loss \cite{wang2019miss}.

All variables are independently encoded into a latent space with $C=32$ channels, which is also the hidden-state dimension of the SOM. All encoder, SOM, LCP updater, and decoder convolutions use $3\times3$ kernels with a stride of 1 and padding of 1. Each $\mathcal{B}/\mathcal{T}/\mathcal{N}$ updater contains three convolutional blocks, each equipped with spectral normalization, 16-group Group Normalization, and ReLU activation.

Before the first unfolded stage, LCPNet explicitly initializes the image-domain variables by setting the background and data-fidelity inputs to the infrared image, i.e., $B^{0}=D$ and $X=D$, while setting the target, noise, and dual variables to zero tensors with the same spatial size as $D$, i.e., $T^{0}=N^{0}=Y^{0}=0$. The SOM hidden state $h^0$ is initialized separately as an all-zero tensor.
The step sizes $\rho$ for updating $\mathcal{B}$, $\mathcal{T}$ and $\mathcal{N}$ are learnable scalar parameters rather than fixed hyperparameters. In each unfolded stage, they are all initialized to $0.1$ and optimized jointly with the rest of the network.

\subsection{State-of-the-Art Comparison}

To provide a comprehensive evaluation, we compare LCPNet with representative IRSTD methods on NUDT-SIRST \cite{li2022dense}, IRSTD-1K \cite{zhang2022isnet}, SIRST \cite{dai2021asymmetric}, and SIRST-Aug \cite{zhang2023attention} under the same evaluation protocol described above. The compared methods cover four categories: HVS-based methods, optimization-based methods, deep learning networks, and recent deep unfolding models. This selection allows us to examine detection accuracy, false-alarm suppression, visual robustness, and the efficiency-accuracy tradeoff across different model families.

\subsubsection{Quantitative Comparison}

Table~\ref{table:sota} shows that LCPNet provides the most consistent quantitative performance among the compared model families. On the two larger benchmarks e.g. NUDT-SIRST \cite{li2022dense}, IRSTD-1K \cite{zhang2022isnet}, LCPNet-6 obtains the best performance, with IoU/$\text{F}_1$ of $95.77\%$/$97.84\%$ and $65.95\%$/$79.48\%$, respectively. More importantly for IRSTD, where small bright clutter easily causes false alarms, the LCPNet variants achieve the lowest $\text{F}_a$ on all four benchmarks, indicating that the proposed decomposition is not merely more accurate in overlap but also more conservative in suppressing background responses.

This behavior is closely related to the design of LCPNet. The latent-space unfolding gives each stage a higher-dimensional representation for separating target and background structures, while the proximal surrogate and shared memory stabilize the update trajectory across stages. On SIRST, LCPNet-6 achieves competitive segmentation accuracy with a low $\text{F}_a$ of $0.89$, although RPCANet++ obtains higher IoU, $\text{F}_1$, and $\text{P}_d$.
On SIRST-Aug, LCPNet-6 achieves the highest IoU/$\text{F}_1$ of $73.68\%$/$84.85\%$ among the compared methods, together with a low $\text{F}_a$ of $2.24$.
Nevertheless, its $\text{P}_d$ remains below that of several baselines on both datasets, highlighting a tradeoff between target detection and false-alarm suppression.

\begin{figure*}[t]
    \centering
    \captionsetup[subfloat]{font=footnotesize,labelfont=footnotesize}
    \subfloat[NUDT-SIRST]{
        \includegraphics[width=0.235\linewidth]{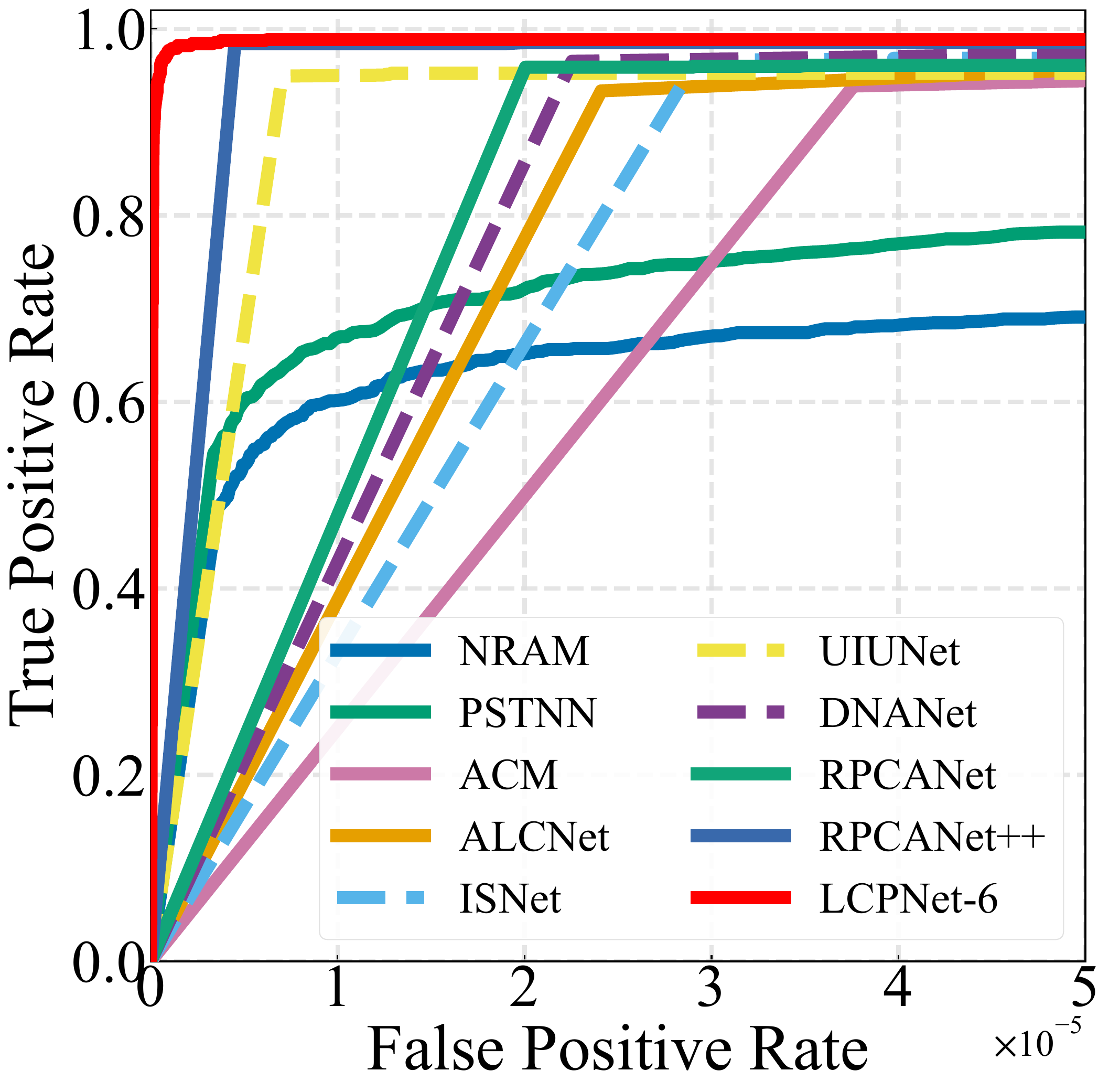}
        \label{fig:roc_nudt_sirst}
    }
    \hfill
    \subfloat[IRSTD-1k]{
        \includegraphics[width=0.235\linewidth]{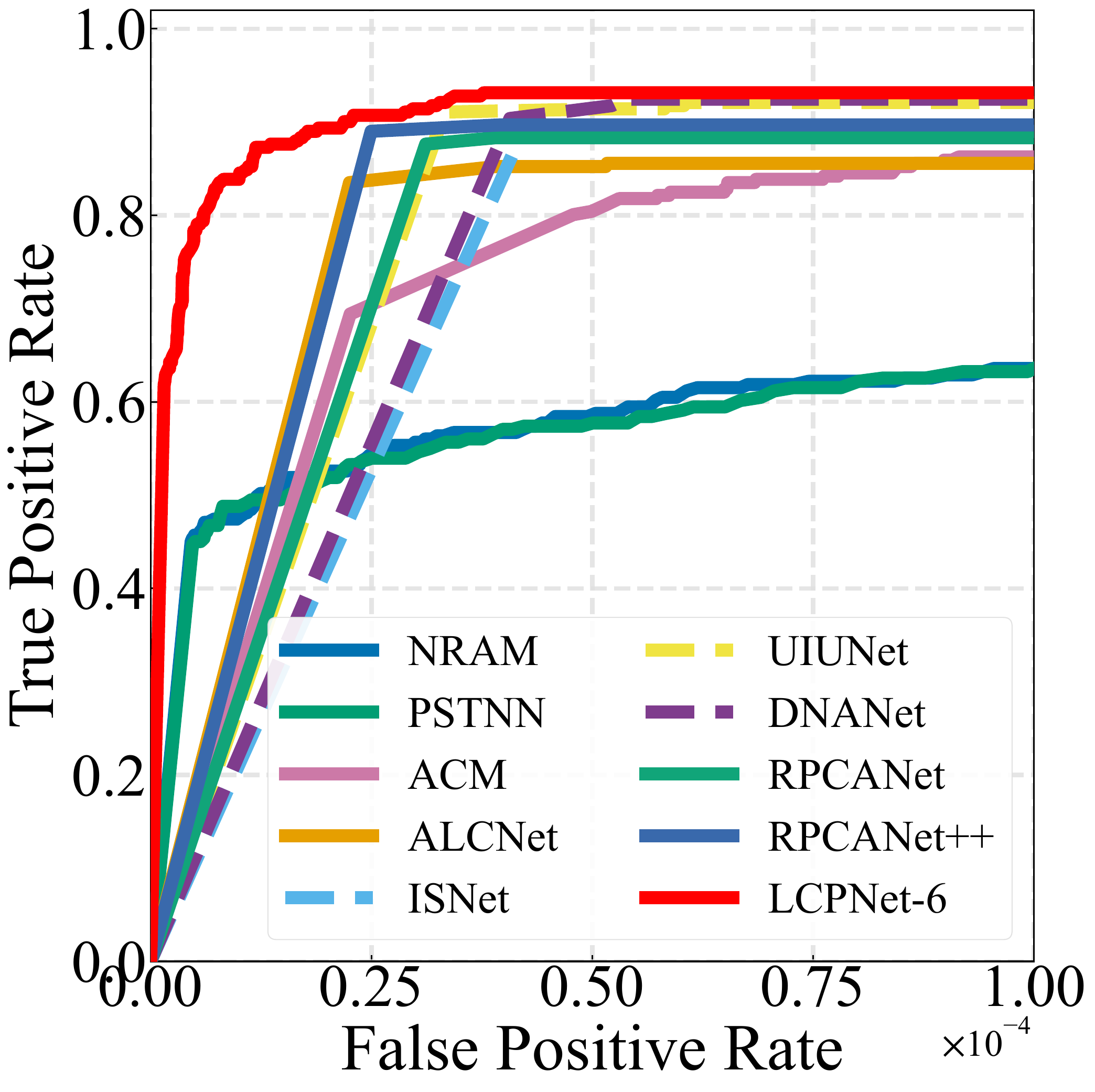}
        \label{fig:roc_irstd1k}
    }
    \hfill
    \subfloat[SIRST]{
        \includegraphics[width=0.235\linewidth]{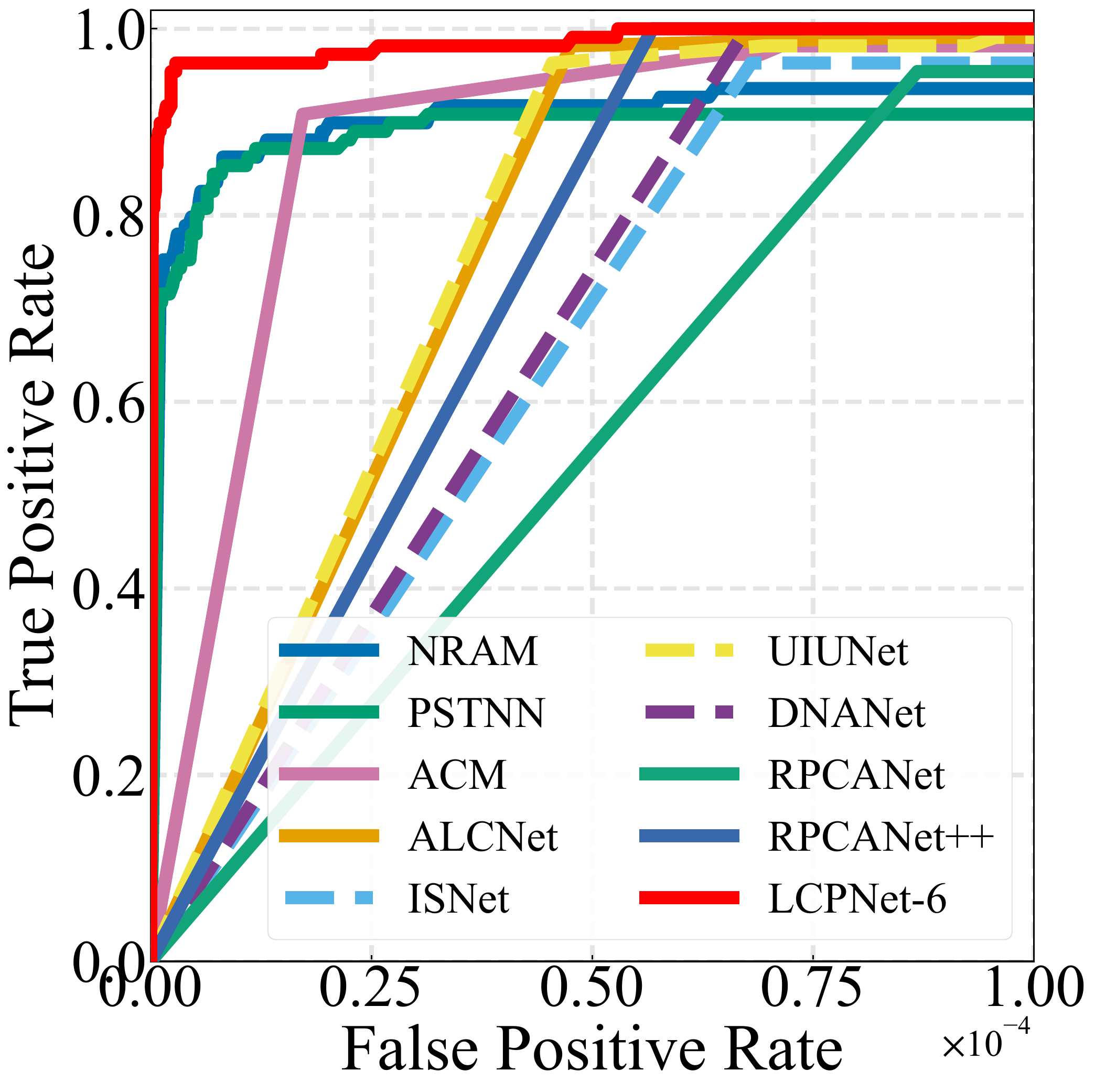}
        \label{fig:roc_sirst}
    }
    \hfill
    \subfloat[SIRST-Aug]{
        \includegraphics[width=0.235\linewidth]{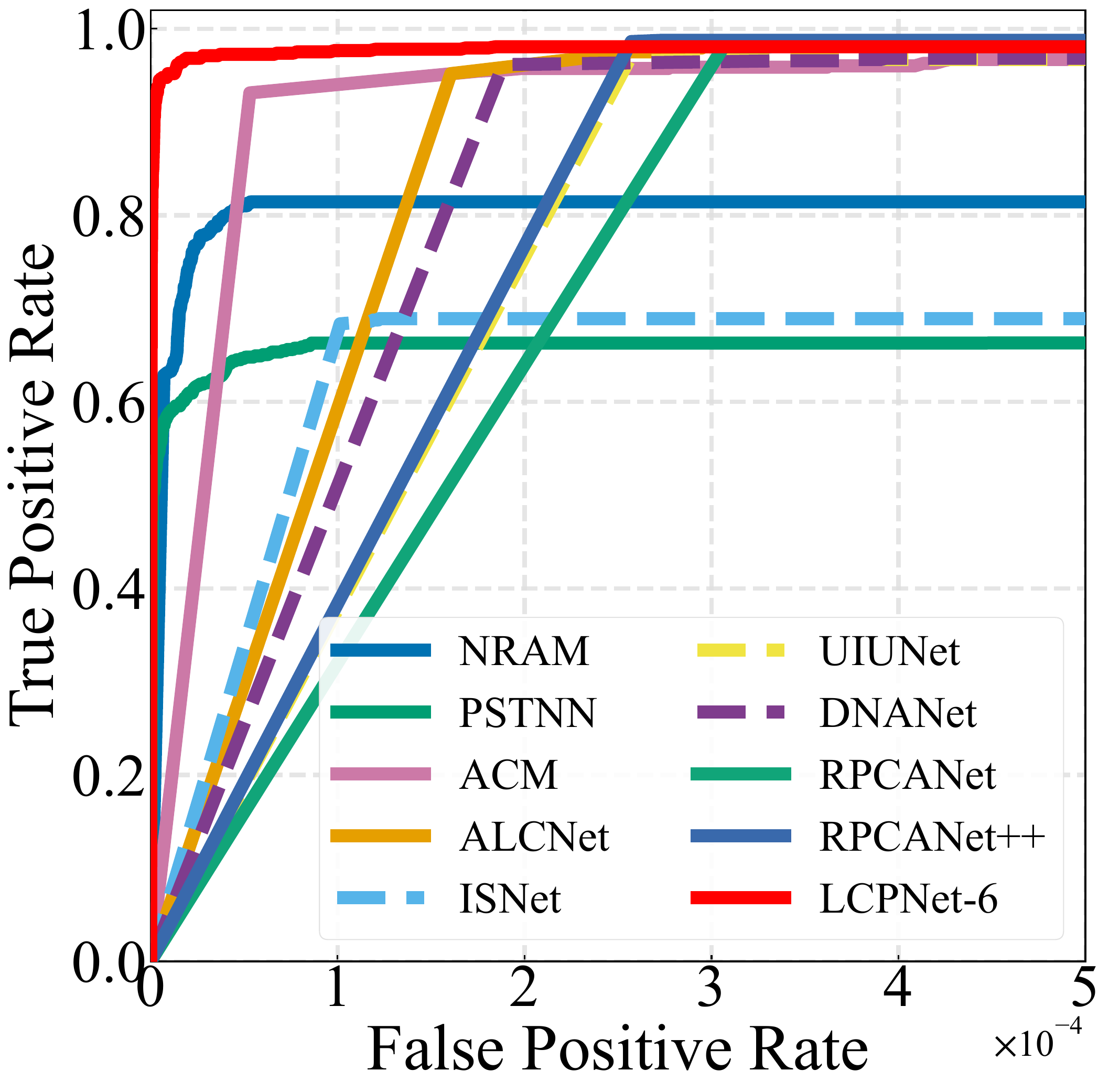}
        \label{fig:roc_sirstaug}
    }
    \caption{ROC curves of different state-of-the-art methods across four infrared small target detection datasets. Our method is represented by the red line.}
    \label{fig:four_rocs}
\end{figure*}

The ROC results in Fig.~\ref{fig:four_rocs} provide the same evidence from a threshold-independent view: after zooming into the high-sensitivity region, the red LCPNet curve stays closest to the upper-left corner. Consistently, Table~\ref{table:auc} reports the best AUC on three benchmarks, where LCPNet-6 reaches $0.9941$, $0.9656$, and $0.999998$, and remains competitive on the augmented one with an AUC of $0.9803$. Overall, these results suggest that the proposed latent space unfolding strategy improves not only peak segmentation metrics but also the robustness of detection decisions across thresholds.

\begin{table}[!t]
    \centering
    \caption{AUC comparison of different methods on four infrared small target detection datasets. Higher values are marked as \textbf{bold} and indicate better detection performance.}
    \label{table:auc}
    \renewcommand\arraystretch{1.2}
    \setlength{\tabcolsep}{1.2mm}

    \resizebox{\linewidth}{!}{
    \begin{tabular}{lcccc}
        \TableOuterRule
        Method & NUDT-SIRST & IRSTD-1K & SIRST & SIRST-Aug \\
        \TableInnerRule

        NRAM \cite{zhang2018infrared} & 0.7403 & 0.7777 & 0.936405 & 0.7794 \\
        \SotaAltRow PSTNN \cite{dai2017reweighted} & 0.8968 & 0.8114 & 0.892235 & 0.5069 \\
        ACM \cite{dai2021asymmetric} & 0.9867 & 0.9363 & 0.990813 & 0.9834 \\
        \SotaAltRow ALCNet \cite{dai2021attentional} & 0.9909 & 0.9281 & 0.999975 & 0.9909 \\
        ISNet \cite{zhang2022isnet} & 0.9861 & 0.9446 & 0.981616 & 0.8436 \\
        \SotaAltRow UIUNet \cite{wu2022uiu} & 0.9761 & 0.9604 & 0.995389 & 0.9833 \\
        DNANet \cite{li2022dense} & 0.9909 & 0.9655 & 0.999966 & 0.9840 \\

        \SotaAltRow RPCANet \cite{wu2024rpcanet} & 0.9804 & 0.9415 & 0.961126 & 0.9909 \\
         RPCANet++ \cite{wu2025rpcanet++} & 0.9925 & 0.9484 & 0.999972 & \textbf{0.9937} \\
        \SotaOursRow LCPNet-6 & \textbf{0.9941} & \textbf{0.9656} & \textbf{0.999998} & 0.9803 \\
        \TableOuterRule
    \end{tabular}
    }
\end{table}

\subsubsection{Qualitative Visual Comparison}

\begin{figure*}[!t]
  \centering
   \includegraphics[width=0.99\linewidth]{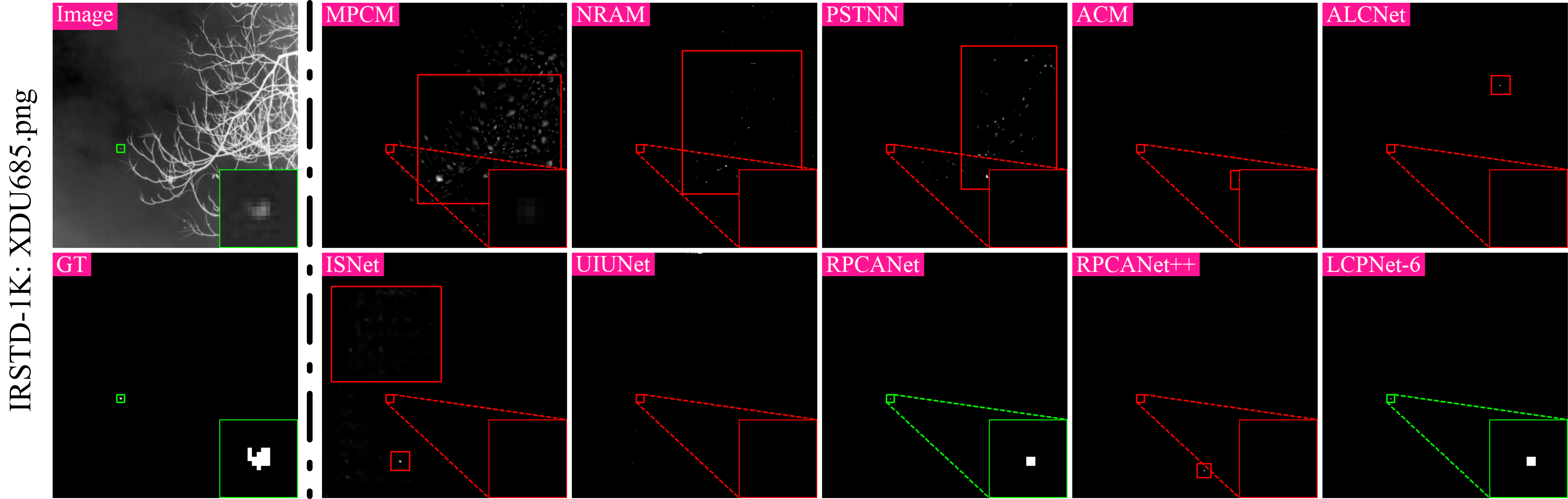}
  \caption{Qualitative comparison on image \textit{XDU685}. Green boxes denote correct detections, while red boxes denote false alarms or missed targets. Best view in color.}
  \label{fig:irstd-1k-xdu685}
\end{figure*}

\begin{figure*}[!t]
  \centering
   \includegraphics[width=0.99\linewidth]{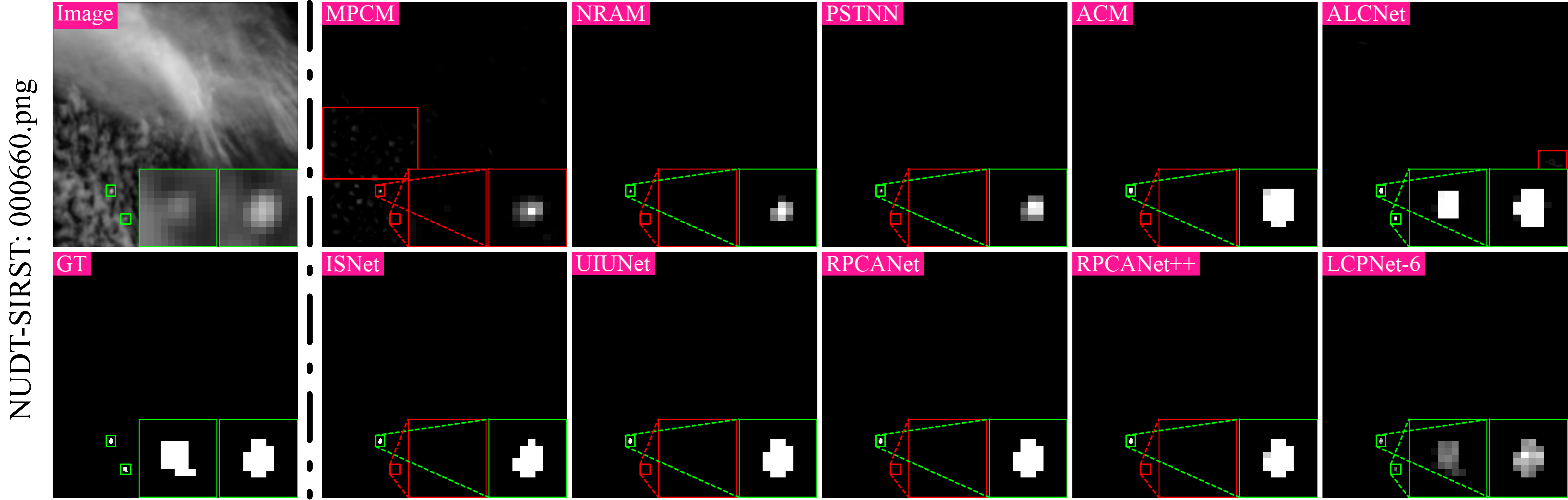}
  \caption{Qualitative comparison on image \textit{000660}. Green boxes denote correct detections, while red boxes denote false alarms or missed targets. Best view in color.}
  \label{fig:nudt-sirst-000660}
\end{figure*}

\begin{figure*}[!t]
  \centering
   \includegraphics[width=0.99\linewidth]{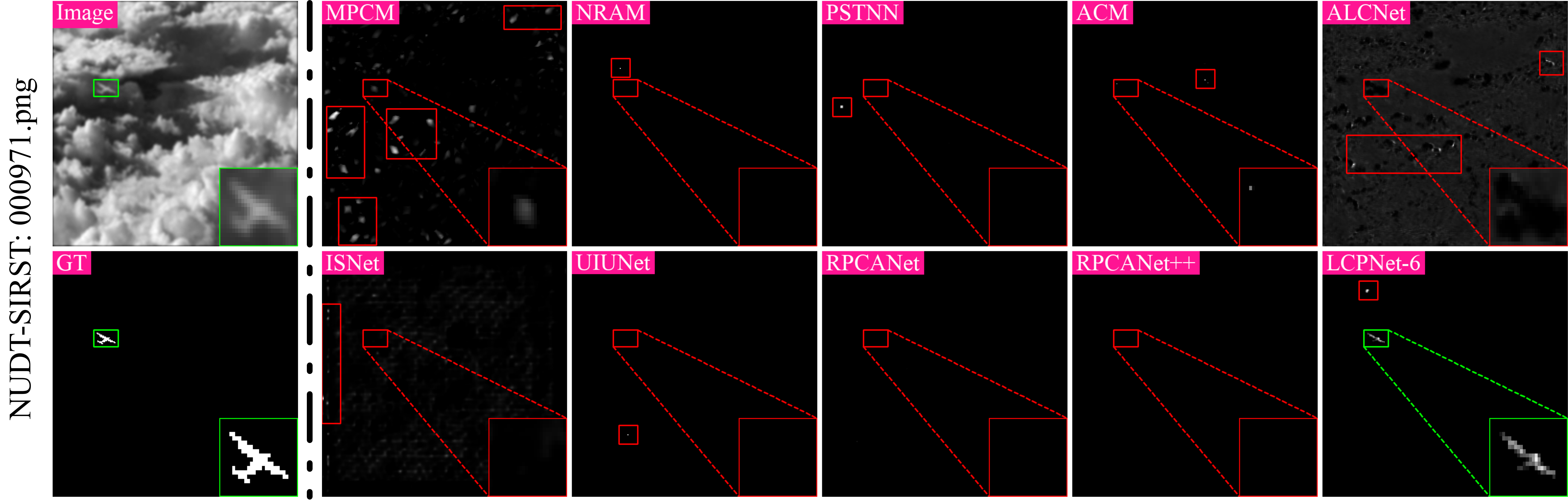}
  \caption{Qualitative comparison on image \textit{000971}. Green boxes denote correct detections, while red boxes denote false alarms or missed targets. Best view in color.}
  \label{fig:nudt-sirst-000971} 
\end{figure*}

\begin{figure*}[!t]
  \centering
   \includegraphics[width=0.99\linewidth]{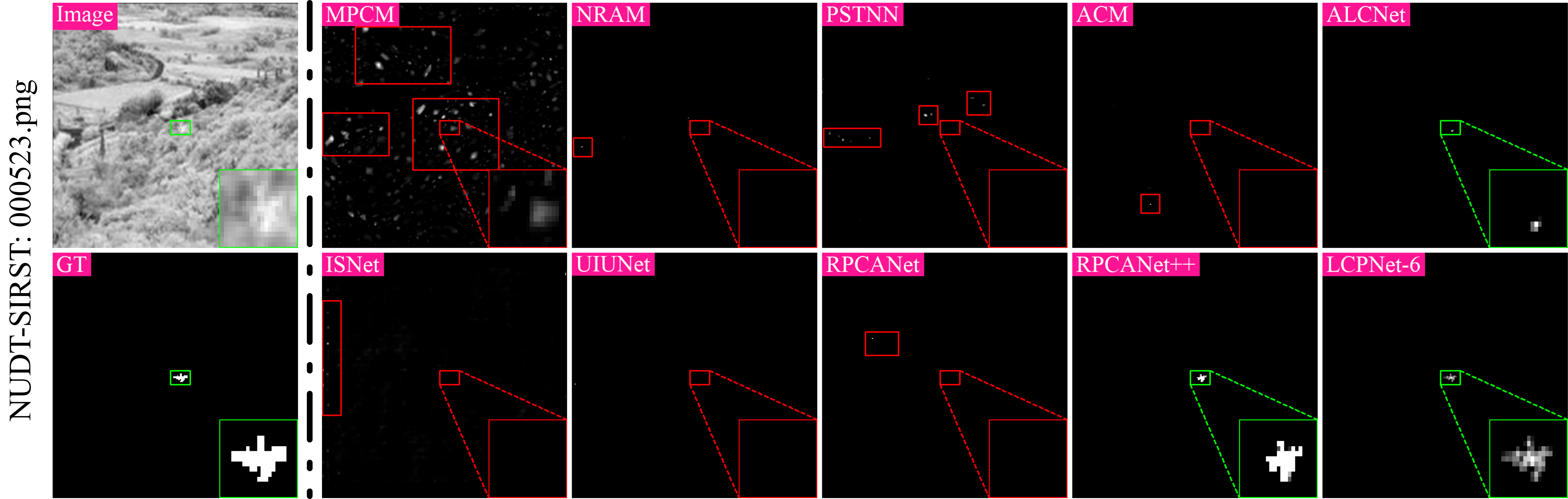}
  \caption{Qualitative comparison on image \textit{000523}. Green boxes denote correct detections, while red boxes denote false alarms or missed targets. Best view in color.}
  \label{fig:nudt-sirst-000523}
\end{figure*}

Fig.~\ref{fig:irstd-1k-xdu685}-Fig.~\ref{fig:nudt-sirst-000523} provide representative visual comparisons, where green boxes denote correct detections and red boxes denote false alarms or missed targets. These examples cover several typical difficult cases, including tree-branch clutter, multiple adjacent targets, cloud background structures, and complex terrain textures. 

In Fig.~\ref{fig:irstd-1k-xdu685}, traditional optimization-based methods such as MPCM, NRAM, and PSTNN leave many clutter responses, while only RPCANet and LCPNet-6 correctly detect the true target. This result suggests that unfolding-based decomposition is more robust than hand-crafted priors when the background contains structured bright interference.
The advantage of LCPNet becomes clearer in Fig.~\ref{fig:nudt-sirst-000660} with multiple targets. Most competing methods either miss one target or mix the target response with nearby background, whereas LCPNet-6 completely detects both targets. 
For the cloud and terrain examples in Fig.~\ref{fig:nudt-sirst-000971} and Fig.~\ref{fig:nudt-sirst-000523}, many baselines are distracted by background textures and generate multiple red-box responses. LCPNet-6 localizes the real targets with more compact responses and fewer clutter activations, although a small residual false alarm remains in the cloud scene. Overall, the visual results support the quantitative findings: LCPNet improves practical detection reliability by better separating small targets from structured background clutter.

To further inspect the feature-level behavior, Fig.~\ref{fig:heatmap_1} visualizes the last-layer feature maps of different methods using the most informative PCA component. Across sky and ground background scenes, as well as single-target, multi-target, and heavy-clutter cases, LCPNet produces clearer target-focused activations than the competing methods. The target responses are more concentrated around true small objects, while background structures are less likely to dominate the heatmaps. 
This suggests that the proposed latent space unfolding process provides more reliable feature evidence for subsequent target detection and segmentation.

%


\subsubsection{Computational Efficiency Analysis}

Beyond detection accuracy, Table~\ref{table:sota} also reports the model parameters, FLOPs, and CPU/GPU inference time. 
HVS-based and optimization-based methods do not involve GPU inference and are therefore evaluated only on the CPU. Their runtimes vary substantially because they rely on different hand-crafted operations or iterative solvers.
For a consistent latency comparison, all GPU-evaluated deep models were tested on an NVIDIA GeForce RTX 3090 using input images with a spatial resolution of $256\times256$ and a batch size of 8. Inference was performed using single-precision floating-point computation (FP32), with automatic mixed precision disabled. Before timing, we ran 50 warm-up iterations to eliminate initialization overhead and stabilize GPU execution. We then averaged the inference latency over 200 measured forward passes. CUDA synchronization was performed after the warm-up stage and after each timed forward pass to ensure that all asynchronous GPU operations were completed before the elapsed time was recorded.

Deep learning-based methods improve inference efficiency by replacing iterative solvers with feed-forward networks, but their cost differs substantially across architectures. Lightweight models such as ACM and ALCNet are fast, whereas larger context aggregation or nested U-shaped designs introduce noticeably higher FLOPs and latency. However, these faster non-unfolding networks usually trade interpretability and often show weaker robustness in the quantitative and visual comparisons above.

Among deep unfolding-based methods, LCPNet provides a practical balance between accuracy, model size, and runtime. Compared with RPCANet++, LCPNet-4 uses fewer parameters and substantially lower FLOPs, while also running faster on GPU. LCPNet-6 increases the unfolding depth to improve accuracy but still keeps the runtime below heavier unfolding baselines such as RPCANet++. Therefore, LCPNet does not achieve its performance by simply enlarging the network. Its efficiency mainly comes from performing compact latent-domain updates with shared optimization memory, yielding a favorable accuracy-efficiency tradeoff for infrared small target detection.

\subsection{Ablation Studies}

To verify the effectiveness of each component in LCPNet, we conduct a series of controlled ablation studies by modifying one design factor at a time while keeping the remaining settings unchanged.
Together, these studies provide an overall validation of the main architectural choices and their impact on detection performance.

\subsubsection{Effect of Execution Domain}
We first examine whether the unfolding process should be executed in the image domain or in the latent domain. In Table~\ref{table:ablation}, variants (b) and (c) use the same surrogate-style solver and batch-based updater, so their comparison isolates the effect of the execution domain. Moving the updates from $\mathbf{X}$ to $\mathcal{X}$ improves IoU by $0.72$, $\text{F}_1$ by $0.39$, and $\text{P}_d$ by $0.84$, indicating that the latent variables provide a more informative representation for preserving weak target responses. However, the $\text{F}_a$ value is not reduced in this setting, which suggests that latent execution alone is insufficient when the updater is weakly constrained.

The benefit becomes clearer when the updater is equipped with stronger regularization. Under the same solver and regularized updater setting, variants (e) and (f) show that latent-domain execution brings gains of $2.83$ in IoU, $1.49$ in $\text{F}_1$, and $1.06$ in $\text{P}_d$, while reducing $\text{F}_a$ from $0.080$ to $0.025$. This result indicates that the expanded latent representation can better separate target and background components when the update process is stable enough to exploit it. Overall, these two comparisons show that latent-domain unfolding is more effective than directly updating image-domain variables.

\subsubsection{Effect of Solver Design}

We evaluate the solver design used to generate the stage-wise update direction. The R-style design, adopted by the RPCANet series \cite{wu2024rpcanet,wu2025rpcanet++,xiong2025drpca}, directly reconstructs the next background variable from the current decomposition residual through a learned correction:
\begin{equation}
    \mathbf{B}^k = \mathbf{D}^{k-1} - \mathbf{T}^{k-1} + \mathcal{F}(\mathbf{D}^{k-1} - \mathbf{T}^{k-1}).
\end{equation}

This form is simple and effective, but the update of $\mathbf{B}^k$ does not explicitly use $\mathbf{B}^{k-1}$. Therefore, the background trajectory is mainly regenerated from the residual at each stage, which weakens the continuity of the unfolding process.

In contrast, the proposed LCP solver design in Eq.~\ref{eq:lcp_solver} keeps the proximal surrogate form while removing the SOM-related state input for a fair comparison.
This design updates current latent variable from its previous state and a decomposition-consistency residual, making the learned direction closer to a proximal descent step rather than an indirect residual mapping.

The comparison between variants (a) and (b) in Table~\ref{table:ablation} verifies this effect under the same image-domain and batch-based updater setting. Replacing the R-style solver with the S-style solver brings gains of $1.27$ in IoU and $0.71$ in $\text{F}_1$, while lowering $\text{F}_a$ by $2.72$. Although $\text{P}_d$ decreases by $3.38$, the much lower false alarm rate and higher overlap metrics indicate that the S-style solver suppresses background responses more effectively. Overall, the proximal surrogate formulation provides a more stable and discriminative update rule for the unfolding decomposition.

\subsubsection{Effect of Updater Regularization}

We further analyze the regularization strategy inside the updater. In Table~\ref{table:ablation}, variants (c), (d), and (f) keep the latent execution domain and S-style solver unchanged, and only vary the updater regularization.

Replacing BN with GN gives the main improvement. From variant (c) to (d), IoU, $\text{F}_1$, and $\text{P}_d$ increase by $4.29$, $2.29$, and $2.65$, respectively, while $\text{F}_a$ is reduced by $0.260$. This large gain is consistent with the characteristics of IRSTD, where infrared scenes often have heterogeneous background statistics and small targets occupy only a tiny portion. A batch-dependent updater can therefore be sensitive to mixed scene statistics, whereas GN normalizes features within each sample and provides a more stable update direction.

Adding spectral normalization on top of GN brings a smaller but consistent gain. From variant (d) to (f), IoU and $\text{F}_1$ further increase by $0.11$ and $0.06$, $\text{P}_d$ improves by $0.32$, and $\text{F}_a$ is further reduced by $0.104$. Overall, GN is the primary factor for improving updater stability, and SN further regularizes the unfolding update process, leading to more reliable target-background separation.

\begin{table}[!t]
    \centering
    \caption{Ablation study of execution domain, solver design, and updater regularization in terms of IoU(\%), $\text{F}_1$(\%), $\text{P}_d$(\%) and $\text{F}_a(10^{-5})$ on NUDT-SIRST \cite{li2022dense}. $\mathbb{B}$, $\mathbb{G}$ and $\mathbb{S}$ represent batch normalization, group normalization, and spectral normalization, respectively.}
    \label{table:ablation}
    \renewcommand\arraystretch{1.3}
    \setlength{\tabcolsep}{1.3mm}
    \resizebox{\linewidth}{!}{
    \begin{tabular}{cccc|cccc}
        \TableOuterRule
        \multirow{2}{*}{Index} & \multirow{2}{*}{Domain} & Solver & Updater & \multicolumn{4}{c}{NUDT-SIRST} \\
        && Design & Regular & IoU$\uparrow$ & $\text{F}_1$$\uparrow$ & $\text{P}_d$$\uparrow$ & $\text{F}_a$$\downarrow$ \\
        \TableInnerRule

        (a) & Image & R-style & $\mathbb{B}$ & 89.31 & 94.35 & 97.14 & 2.87 \\
        \SotaAltRow (b) & Image & LCP & $\mathbb{B}$ & 90.58 & 95.06 & 93.76 & 0.152 \\
        (c) & Latent & LCP & $\mathbb{B}$ & 91.30 & 95.45 & 94.60 & 0.389 \\
        \SotaAltRow (d) & Latent & LCP & $\mathbb{G}$ & 95.59 & 97.74 & 97.25 & 0.129 \\
        (e) & Image & LCP & $\mathbb{G}+\mathbb{S}$ & 92.87 & 96.31 & 96.51 & 0.080 \\
        \SotaOursRow (f) & Latent & LCP & $\mathbb{G}+\mathbb{S}$ & 95.70 & 97.80 & 97.57 & 0.025 \\


        \TableOuterRule
    \end{tabular}
    }
\end{table}

\subsubsection{Effect of Memory Type}

We then evaluate how historical information is represented across unfolding stages. Table~\ref{table:ablation_memory} compares four settings, including no memory, simple concatenation, ConvLSTM-style recurrent memory used in RPCANet++ \cite{wu2025rpcanet++}, and the proposed SOM. Compared with no memory, simple concatenation brings only a marginal IoU gain of $0.21$ on NUDT-SIRST and shows no clear improvement on IRSTD-1K, indicating that static feature aggregation is insufficient for modeling the stage-wise optimization history. ConvLSTM gives more consistent gains, improving IoU by $0.31$ and $0.37$ on the two benchmarks and reducing $\text{F}_a$ on both datasets. This result confirms that recurrent memory is useful for preserving cross-stage information.

SOM further improves segmentation quality by treating the decomposition variables as a coupled optimization state rather than storing branch-specific memory. Compared with ConvLSTM, SOM increases IoU/$\text{F}_1$ by $0.27$/$0.14$ on NUDT-SIRST and $0.43$/$0.31$ on IRSTD-1K. It also improves $\text{P}_d$ on NUDT-SIRST by $0.32$ and reduces $\text{F}_a$ by $0.021$, although its $\text{F}_a$ on IRSTD-1K is not the lowest. Overall, SOM provides stronger overlap accuracy and more stable target preservation, showing that a shared optimization memory is more effective than others.

\begin{table}[!t]
    \centering
    \caption{Ablation study of memory type in terms of IoU(\%), $\text{F}_1$(\%), $\text{P}_d$(\%) and $\text{F}_a(10^{-5})$ on NUDT-SIRST \cite{li2022dense} and IRSTD-1K \cite{zhang2022isnet}.}
    \label{table:ablation_memory}
    \renewcommand\arraystretch{1.3}
    \setlength{\tabcolsep}{0.5mm}
    \resizebox{\linewidth}{!}{
    \begin{tabular}{c|cccc|cccc}
        \TableOuterRule
        \multirow{2}{*}{Memory} & \multicolumn{4}{c|}{NUDT-SIRST} & \multicolumn{4}{c}{IRSTD-1K} \\
        & IoU$\uparrow$ & $\text{F}_1$$\uparrow$ & $\text{P}_d$$\uparrow$ & $\text{F}_a$$\downarrow$ & IoU$\uparrow$ & $\text{F}_1$$\uparrow$ & $\text{P}_d$$\uparrow$ & $\text{F}_a$$\downarrow$ \\
        \TableInnerRule

        None & 95.12 & 97.50 & 96.51 & 0.074 & 64.35 & 78.31 & 86.60 & 1.45 \\
        \SotaAltRow Cat & 95.33 & 97.61 & 96.40 & 0.051 & 64.31 & 78.28 & 85.57 & 1.41 \\
        ConvLSTM & 95.43 & 97.66 & 97.25 & 0.046 & 64.72 & 78.58 & 86.94 & 1.38 \\
        \SotaOursRow SOM & 95.70 & 97.80 & 97.57 & 0.025 & 65.15 & 78.89 & 86.60 & 1.58 \\
        
        \TableOuterRule
    \end{tabular}
    }
\end{table}

\subsubsection{Effect of Stage Number}

Finally, we study the influence of the unfolding stage number in Table~\ref{table:ablation_stage}. Increasing the number of stages generally improves the detection quality because more decomposition updates allow the latent variables and the shared state to be refined progressively. For example, increasing the depth from $K=2$ to $K=4$ improves IoU/$\text{F}_1$ by $1.43$/$0.75$ on NUDT-SIRST and $0.87$/$0.64$ on IRSTD-1K, while $\text{P}_d$ also increases on both benchmarks.

The improvement becomes less regular when the network is further deepened. The setting $K=5$ gives a notably high result on NUDT-SIRST, but this behavior is more like an isolated fluctuation under the current evaluation rather than the main trend. We therefore do not take it as the primary result. Overall, a moderate-to-deep unfolding depth provides better representation refinement, while excessively increasing the number of stages brings smaller and less stable gains.

\begin{table}[!t]
    \centering
    \caption{Ablation study of the unfolding stage number in terms of IoU(\%), $\text{F}_1$(\%), $\text{P}_d$(\%) and $\text{F}_a(10^{-5})$ on NUDT-SIRST \cite{li2022dense} and IRSTD-1K \cite{zhang2022isnet}.}
    \label{table:ablation_stage}
    \renewcommand\arraystretch{1.3}
    \setlength{\tabcolsep}{0.8mm}
    \resizebox{\linewidth}{!}{
    \begin{tabular}{c|cccc|cccc}
        \TableOuterRule
        \multirow{2}{*}{Stages} & \multicolumn{4}{c|}{NUDT-SIRST} & \multicolumn{4}{c}{IRSTD-1K} \\
        & IoU$\uparrow$ & $\text{F}_1$$\uparrow$ & $\text{P}_d$$\uparrow$ & $\text{F}_a$$\downarrow$ & IoU$\uparrow$ & $\text{F}_1$$\uparrow$ & $\text{P}_d$$\uparrow$ & $\text{F}_a$$\downarrow$ \\
        \TableInnerRule

        \SotaAltRow K=2 & 94.27 & 97.05 & 95.03 & 0.046 & 64.28 & 78.25 & 82.47 & 1.12 \\
        K=3 & 95.16 & 97.52 & 97.14 & 0.011 & 64.36 & 78.31 & 84.88 & 1.29 \\
        \SotaAltRow K=4 & 95.70 & 97.80 & 97.57 & 0.025 & 65.15 & 78.89 & 86.60 & 1.58 \\
        K=5 & 96.11 & 98.02 & 98.52 & 0.028 & 65.85 & 79.41 & 87.63 & 1.53 \\
        \SotaAltRow K=6 & 95.77 & 97.84 & 97.35 & 0.092 & 65.95 & 79.48 & 87.63 & 1.55 \\
        
        \TableOuterRule
    \end{tabular}
    }
\end{table}

\subsection{Discussion and Analysis}

\subsubsection{Technical Discussion}

LCPNet follows the same model-driven spirit as recent deep unfolding methods, but differs from them in how the decomposition process is represented and updated. Traditional optimization-based methods operate directly in the image domain and rely on hand-crafted low-rank/sparse priors, while RPCANet-style networks unfold an RPCA-like process with learnable modules in the image domain. LCPNet instead lifts the observation and decomposition variables into a latent domain. This transformation does not discard the physical decomposition constraint, but gives each stage a higher-dimensional interface where small target responses, background structures, and noise components can evolve more flexibly.

The solver design also differs from the residual-style update used in RPCANet-related methods. In the R-style formulation, the next background estimate is mainly regenerated from the current residual through a learned correction, and the previous background state is not explicitly involved in the update. LCPNet adopts an S-style proximal surrogate update, where each variable is updated from its previous state and the current decomposition-consistency residual. This makes the learned update closer to a proximal descent step and gives the unfolding trajectory stronger continuity. Moreover, the SOM further separates LCPNet from RPCANet++. The memory module in RPCANet++ mainly serves a specific branch, whereas SOM encodes the coupled state of $\mathcal{B}$, $\mathcal{T}$, $\mathcal{N}$, $\mathcal{Y}$, and $\mathcal{X}$ and feeds the same historical state to all primal-variable updates. Through the gated recurrence, this state forms an unbounded-order effective memory of the whole latent decomposition system rather than a branch-level feature cache.

\subsubsection{Limitations and Future Work}

Although LCPNet achieves a favorable accuracy-efficiency tradeoff in deep-unfolding based methods, its inference time still has room for further reduction. The latent-domain updates and recurrent state propagation introduce additional computation compared with very lightweight feed-forward detectors, especially when more unfolding stages are used. 
Moreover, LCPNet is inspired by an explicit optimization formulation, its solver and proximal directions are implemented by learnable neural modules rather than exact analytical operators. Consequently, the current formulation does not provide a strict convergence guarantee, and the effectiveness of the resulting finite-stage network is instead validated empirically through experiments.
Future work can further compress the updater, simplify the state transition, or design adaptive early-exit strategies so that easy samples require fewer stages. We expect these directions to support more efficient and robust infrared small target detectors, especially for real-time deployment on resource-constrained platforms. 

\section{Conclusion}

In this paper, we introduced LCPNet for infrared small target detection. 
By confirming that low-rank decomposition prior remains valid in latent representations, LCPNet extends unfolding from image domain to latent domain, preserving the physical constraint while avoiding repeated compression of intermediate states. 
Through modeling and detailed derivations on latents, the proposed LCP solver turns indirect residual reconstruction into direct proximal-state evolution and introduces regularized update dynamics for stable small target separation. 
The SOM further upgrades branch-specific memory into a shared optimization state, so historical information can guide all coupled decomposition variables across stages.
Extensive experiments on four public benchmarks demonstrate the effectiveness of LCPNet. 
Quantitative and qualitative results show that LCPNet achieves accurate and robust detection across cluttered scenes and multiple-target cases. Ablation studies progressively demonstrate that these benefits stem from the proposed improvements. Overall, LCPNet provides an effective and interpretable framework for IRSTD.


\bibliographystyle{IEEEtran}
\bibliography{ref}

\clearpage

\appendices
\input{appendix}


 




\vfill

\end{document}

%% file: appendix.tex
\section{Derivation of ADMM Subproblems}
\label{app:admm_subproblem_derivation}

This section derives the background, target, and noise subproblems from the augmented Lagrangian in Eq.~\ref{eq:latent_augmented_lagrangian}. The dual and quadratic penalty terms in Eq.~\ref{eq:latent_augmented_lagrangian} can be merged by completing the square:
\begin{equation}
    \begin{aligned}
        &\left\langle \mathcal{Y},\mathcal{X}-\mathcal{B}-\mathcal{T}-\mathcal{N}\right\rangle+\frac{\mu}{2}\left\|\mathcal{X}-\mathcal{B}-\mathcal{T}-\mathcal{N}\right\|_F^2 \\
        &=\frac{\mu}{2}\left\|\mathcal{X}-\mathcal{B}-\mathcal{T}-\mathcal{N}+\frac{1}{\mu}\mathcal{Y}\right\|_F^2-\frac{1}{2\mu}\left\|\mathcal{Y}\right\|_F^2.
    \end{aligned}
\end{equation}
Since the last term is independent of $\mathcal{B}$, $\mathcal{T}$, and $\mathcal{N}$ during primal-variable updates, it can be absorbed into a constant $C$. Therefore, the equivalent primal objective is
\begin{equation}
    \begin{aligned}
        \mathcal{L}_{\mu}(\mathcal{B},\mathcal{T},\mathcal{N},&\mathcal{Y})=\ \mathcal{R}_{\mathcal{B}}(\mathcal{B})+\mathcal{R}_{\mathcal{T}}(\mathcal{T})+\mathcal{R}_{\mathcal{N}}(\mathcal{N};\sigma) \\
        &+\frac{\mu}{2}\left\|\mathcal{X}-\mathcal{B}-\mathcal{T}-\mathcal{N}+\frac{1}{\mu}\mathcal{Y}\right\|_F^2+C.
    \end{aligned}
\end{equation}

ADMM then updates one latent component while fixing the others. For the background component:
\begin{equation}
    \begin{aligned}
        \mathcal{B}^{k}=&\arg\min_{\mathcal{B}}\ \mathcal{R}_{\mathcal{B}}(\mathcal{B}) \\
        &+\frac{\mu}{2}\left\|\mathcal{X}-\mathcal{B}-\mathcal{T}^{k-1}-\mathcal{N}^{k-1}+\frac{1}{\mu}\mathcal{Y}^{k-1}\right\|_F^2.
    \end{aligned}
\end{equation}
For the target component:
\begin{equation}
    \begin{aligned}
        \mathcal{T}^{k}=&\arg\min_{\mathcal{T}}\ \mathcal{R}_{\mathcal{T}}(\mathcal{T}) \\
        &+\frac{\mu}{2}\left\|\mathcal{X}-\mathcal{B}^{k}-\mathcal{T}-\mathcal{N}^{k-1}+\frac{1}{\mu}\mathcal{Y}^{k-1}\right\|_F^2.
    \end{aligned}
\end{equation}
For the noise component:
\begin{equation}
    \begin{aligned}
        \mathcal{N}^{k}=&\arg\min_{\mathcal{N}}\ \mathcal{R}_{\mathcal{N}}(\mathcal{N};\sigma) \\
        &+\frac{\mu}{2}\left\|\mathcal{X}-\mathcal{B}^{k}-\mathcal{T}^{k}-\mathcal{N}+\frac{1}{\mu}\mathcal{Y}^{k-1}\right\|_F^2.
    \end{aligned}
\end{equation}

These three subproblems have the same latent structure: each branch minimizes an unknown latent regularizer plus a quadratic consistency term. This shared structure motivates the surrogate solver derived in the next section.

After the three primal variables are updated, ADMM updates the latent dual variable by accumulating the decomposition residual:
\begin{equation}
    \mathcal{Y}^{k}=\mathcal{Y}^{k-1}+\mu\left(\mathcal{X}-\mathcal{B}^{k}-\mathcal{T}^{k}-\mathcal{N}^{k}\right).
\end{equation}

\begin{figure*}[!t]
  \centering
   \includegraphics[width=0.9\linewidth]{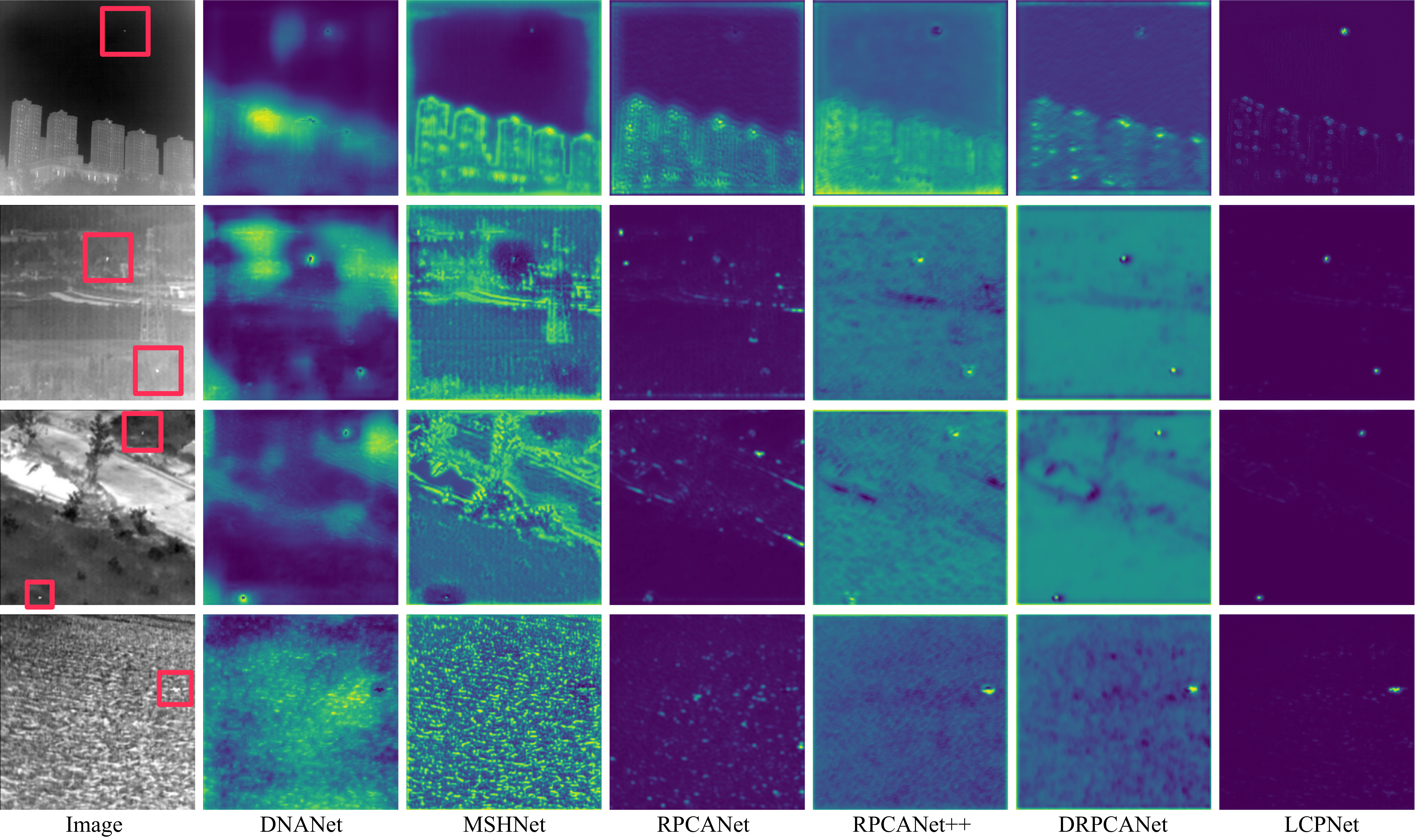}
  \caption{Supplementary on the visualization of last layer feature maps produced by different methods. For each method, the feature map is projected by PCA, and the most informative principal component is selected for visualization. Rows 1 is \textit{XDU67} in IRSTD-1k \cite{zhang2022isnet}. Row 2,3,4 are \textit{000762} \textit{000410} \textit{000589} in NUDT-SIRST \cite{li2022dense}.}
  \label{fig:heatmap_2}
\end{figure*}

\section{Detailed Derivation of the ADMM Surrogate Update}
\label{app:admm_surrogate_derivation}

This section derives the analytical update of the $\mathcal{B}$-subproblem under the Lipschitz-continuous gradient assumption. Starting from the background subproblem in Eq.~\ref{eq:B_proximal_subproblem}, we first define
\begin{equation}
    \mathcal{V}_{B}^{k-1}=\mathcal{X}-\mathcal{T}^{k-1}-\mathcal{N}^{k-1}+\frac{1}{\mu_B}\mathcal{Y}^{k-1}.
\end{equation}
Then the $\mathcal{B}$-subproblem can be written as
\begin{equation}
    \mathcal{B}^{k}=\arg\min_{\mathcal{B}}\ \mathcal{R}_{\mathcal{B}}(\mathcal{B})+\frac{\mu_B}{2}\left\|\mathcal{B}-\mathcal{V}_{B}^{k-1}\right\|_F^2.
\label{app:eq:B_proximal_subproblem}
\end{equation}

Because hand-crafted constraints cannot be perfectly aligned with the physical prior of the latent background, we treat $\mathcal{R}_{\mathcal{B}}(\cdot)$ as an unknown regularizer rather than manually specifying its analytical form. We then use the Taylor upper-bound argument induced by the descent lemma to construct a quadratic majorization. Assume that $\nabla\mathcal{R}_{\mathcal{B}}$ is $L_B$-Lipschitz continuous, i.e.,
\begin{equation}
    \left\|\nabla\mathcal{R}_{\mathcal{B}}(\mathcal{U})-\nabla\mathcal{R}_{\mathcal{B}}(\mathcal{V})\right\|_F\leq L_B\left\|\mathcal{U}-\mathcal{V}\right\|_F.
\label{app:eq:RB_lipschitz}
\end{equation}
Then $\mathcal{R}_{\mathcal{B}}(\mathcal{B})$ satisfies the following upper bound around $\mathcal{B}^{k-1}$:
\begin{equation}
    \begin{aligned}
        \mathcal{R}_{\mathcal{B}}(\mathcal{B})\leq &\ \mathcal{R}_{\mathcal{B}}(\mathcal{B}^{k-1})+\left\langle\nabla\mathcal{R}_{\mathcal{B}}(\mathcal{B}^{k-1}),\mathcal{B}-\mathcal{B}^{k-1}\right\rangle \\
        &+\frac{L_B}{2}\left\|\mathcal{B}-\mathcal{B}^{k-1}\right\|_F^2,\quad \forall \mathcal{B}\in\mathbb{R}^{H\times W\times C}.
    \end{aligned}
\end{equation}
Completing the square gives the equivalent surrogate regularizer
\begin{equation}
    \hat{\mathcal{R}}_{\mathcal{B}}(\mathcal{B},\mathcal{B}^{k-1})=\frac{L_B}{2}\left\|\mathcal{B}-\left(\mathcal{B}^{k-1}-\frac{1}{L_B}\nabla\mathcal{R}_{\mathcal{B}}(\mathcal{B}^{k-1})\right)\right\|_F^2+C_B,
\label{app:eq:B_surrogate_regularizer}
\end{equation}
where $C_B$ is independent of $\mathcal{B}$ and therefore does not affect the minimizer. Substituting the surrogate regularizer in Eq.~\ref{app:eq:B_surrogate_regularizer} into the $\mathcal{B}$-subproblem in Eq.~\ref{app:eq:B_proximal_subproblem} gives
\begin{equation}
    \begin{aligned}
        \hat{\mathcal{J}}_{\mathcal{B}}(\mathcal{B})=&\ \frac{L_B}{2}\left\|\mathcal{B}-\left(\mathcal{B}^{k-1}-\frac{1}{L_B}\nabla\mathcal{R}_{\mathcal{B}}(\mathcal{B}^{k-1})\right)\right\|_F^2 \\
        &+\frac{\mu_B}{2}\left\|\mathcal{B}-\mathcal{V}_{B}^{k-1}\right\|_F^2.
    \end{aligned}
\end{equation}
Since $\hat{\mathcal{J}}_{\mathcal{B}}(\mathcal{B})$ is the sum of two quadratic terms, we differentiate it with respect to $\mathcal{B}$:
\begin{equation}
    \nabla\hat{\mathcal{J}}_{\mathcal{B}}(\mathcal{B})=L_B\left(\mathcal{B}-\mathcal{B}^{k-1}+\frac{1}{L_B}\nabla\mathcal{R}_{\mathcal{B}}(\mathcal{B}^{k-1})\right)+\mu_B\left(\mathcal{B}-\mathcal{V}_{B}^{k-1}\right).
\end{equation}
Setting $\nabla\hat{\mathcal{J}}_{\mathcal{B}}(\mathcal{B})=0$ yields
\begin{equation}
    L_B\left(\mathcal{B}-\mathcal{B}^{k-1}+\frac{1}{L_B}\nabla\mathcal{R}_{\mathcal{B}}(\mathcal{B}^{k-1})\right)+\mu_B\left(\mathcal{B}-\mathcal{V}_{B}^{k-1}\right)=0.
\end{equation}
Rearranging the above equation gives the optimal solution
\begin{equation}
    \mathcal{B}^{k}=\frac{L_B}{L_B+\mu_B}\mathcal{B}^{k-1}+\frac{\mu_B}{L_B+\mu_B}\mathcal{V}_{B}^{k-1}-\frac{1}{L_B+\mu_B}\nabla\mathcal{R}_{\mathcal{B}}(\mathcal{B}^{k-1}).
\end{equation}

\section{Derivation of the High-Order Memory in SOM}
\label{app:sim_memory_expansion}

This section derives the high-order memory form of the Shared Optimization Memory (SOM). At stage $k$, SOM first constructs a joint optimization observation:
\begin{equation}
    o_k=[\mathcal{B}^{k},\mathcal{T}^{k},\mathcal{N}^{k},\mathcal{Y}^{k},\mathcal{X}],
\label{app:eq:sosim_observation}
\end{equation}
where $o_k$ denotes the concatenated latent observation, $\mathcal{B}^{k}$, $\mathcal{T}^{k}$, and $\mathcal{N}^{k}$ are the background, target, and noise variables at stage $k$, $\mathcal{Y}^{k}$ is the latent dual variable, and $\mathcal{X}$ is the lifted infrared image representation. The optimization state cell updates the shared state $h_k$ by
\begin{equation}
    \begin{gathered}
        z_k=\sigma(W_z*[o_k,h_{k-1}]+b_z),\\
        r_k=\sigma(W_r*[o_k,h_{k-1}]+b_r),\\
        \tilde{h}_k=\tanh(W_q*[o_k,r_k\odot h_{k-1}]+b_q),\\
        h_k=(1-z_k)\odot h_{k-1}+z_k\odot\tilde{h}_k.
    \end{gathered}
\label{app:eq:sosim_cell}
\end{equation}

Here $h_k$ is the shared optimization state at stage $k$. The functions $\sigma(\cdot)$ and $\tanh(\cdot)$ denote the sigmoid and hyperbolic tangent activations. Since $z_k$, $r_k$, $\tilde{h}_k$, and $h_k$ have compatible tensor shapes, all products below are element-wise.

The last line of Eq.~\ref{app:eq:sosim_cell} can be viewed as a gated interpolation between the previous state and the current candidate state. To expose its memory structure, we expand the state transition. For stage $k$, we have
\begin{equation}
    h_k=(1-z_k)\odot h_{k-1}+z_k\odot\tilde{h}_k.
\label{app:eq:sosim_step_k}
\end{equation}

The previous state $h_{k-1}$ follows the same recurrence:
\begin{equation}
    h_{k-1}=(1-z_{k-1})\odot h_{k-2}+z_{k-1}\odot\tilde{h}_{k-1}.
\label{app:eq:sosim_step_k_minus_1}
\end{equation}

Substituting Eq.~\ref{app:eq:sosim_step_k_minus_1} into Eq.~\ref{app:eq:sosim_step_k} gives
\begin{equation}
    \begin{aligned}
        h_k
        =&\ (1-z_k)\odot\left[(1-z_{k-1})\odot h_{k-2}+z_{k-1}\odot\tilde{h}_{k-1}\right] \\
        &+z_k\odot\tilde{h}_k \\
        =&\ (1-z_k)\odot(1-z_{k-1})\odot h_{k-2} \\
        &+(1-z_k)\odot z_{k-1}\odot\tilde{h}_{k-1}
        +z_k\odot\tilde{h}_k.
    \end{aligned}
\label{app:eq:sosim_two_step_expansion}
\end{equation}

Similarly, expanding $h_{k-2}$ and substituting it into Eq.~\ref{app:eq:sosim_two_step_expansion} yields
\begin{equation}
    \begin{aligned}
        h_k
        =&\ (1-z_k)\odot(1-z_{k-1})\odot(1-z_{k-2})\odot h_{k-3} \\
        &+(1-z_k)\odot(1-z_{k-1})\odot z_{k-2}\odot\tilde{h}_{k-2} \\
        &+(1-z_k)\odot z_{k-1}\odot\tilde{h}_{k-1}
        +z_k\odot\tilde{h}_k.
    \end{aligned}
\label{app:eq:sosim_three_step_expansion}
\end{equation}

Eq.~\ref{app:eq:sosim_three_step_expansion} shows the pattern of the recursive expansion, each historical candidate $\tilde{h}_t$ is first injected through its own update gate $z_t$, and then propagated to stage $k$ through the retention gates $(1-z_{t+1}),\ldots,(1-z_k)$.

After expanding the recurrence for $m$ steps, where $1\leq m\leq k$, the state can be written as:
\begin{equation}
    \begin{aligned}
        h_k=&
        \left(\prod_{i=k-m+1}^{k}(1-z_i)\right)\odot h_{k-m} \\
        &+\sum_{t=k-m+1}^{k}
        \left[
            z_t\odot\tilde{h}_t\odot
            \prod_{i=t+1}^{k}(1-z_i)
        \right].
    \end{aligned}
\label{app:eq:sosim_m_step_expansion}
\end{equation}

When $t=k$, the product $\prod_{i=k+1}^{k}(1-z_i)$ is an empty product and is defined as an all-one tensor with the same shape as $h_k$, so the last term becomes $z_k\odot\tilde{h}_k$.
Setting $m=k$ expands the recurrence back to the initial state $h_0$:
\begin{equation}
    h_k=\left(\prod_{i=1}^{k}(1-z_i)\right)\odot h_0 + \sum_{t=1}^{k}\left[ z_t\odot\tilde{h}_t\odot \prod_{i=t+1}^{k}(1-z_i) \right].
\label{app:eq:sosim_full_expansion}
\end{equation}

\begin{figure*}[!t]
  \centering
   \includegraphics[width=0.99\linewidth]{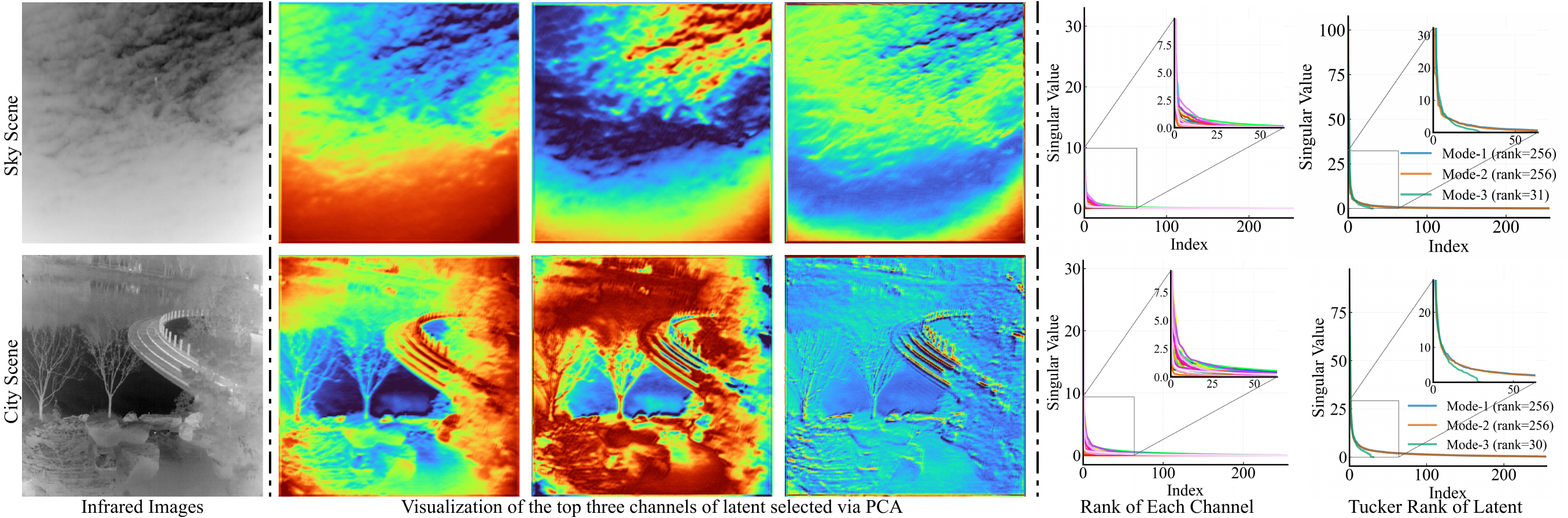}
  \caption{Visualization of the low-rank property in latent domain. From left to right are the original infrared image, a visualized top three channels of the latent selected via PCA, rank of each channel and Tucker rank analysis of the latent. It can be observed that the latent exhibits distinct low-rank property in both visual visualization and rank analysis. The images include sky scene (Top \textit{XDU145} in IRSTD-1k \cite{zhang2022isnet}) and ground scene (bottom \textit{XDU202}).}
  \label{fig:low-rank-2}
\end{figure*}

\begin{figure*}[!t]
  \centering
   \includegraphics[width=0.99\linewidth]{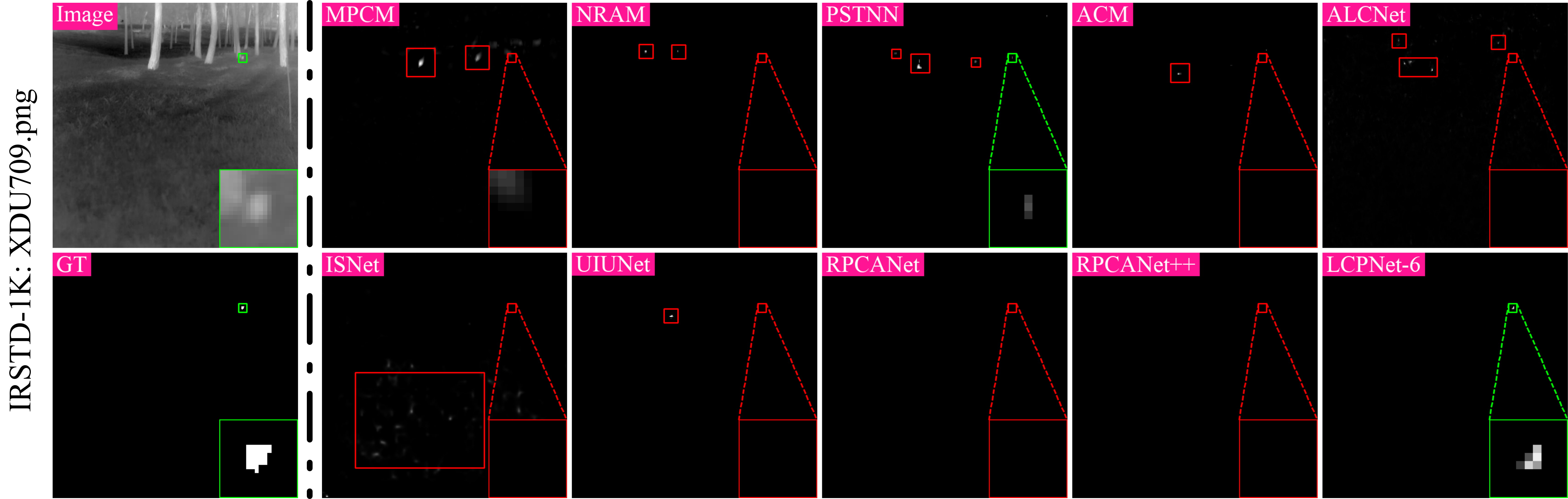}
  \caption{Qualitative comparison on image \textit{XDU709}. Green boxes denote correct detections, while red boxes denote false alarms or missed targets. Best view in color.}
  \label{fig:irstd-1k-xdu709}
\end{figure*}

\begin{figure*}[!t]
  \centering
   \includegraphics[width=0.99\linewidth]{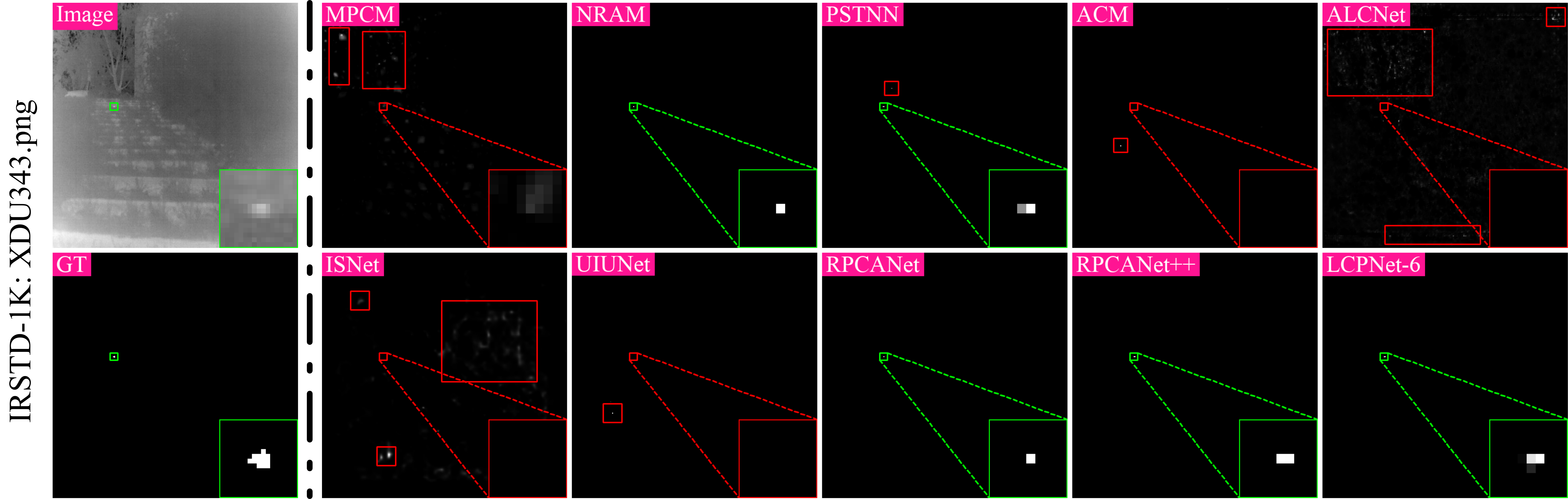}
  \caption{Qualitative comparison on image \textit{XDU343}. Green boxes denote correct detections, while red boxes denote false alarms or missed targets. Best view in color.}
  \label{fig:irstd-1k-xdu343}
\end{figure*}

\begin{figure*}[!t]
  \centering
   \includegraphics[width=0.99\linewidth]{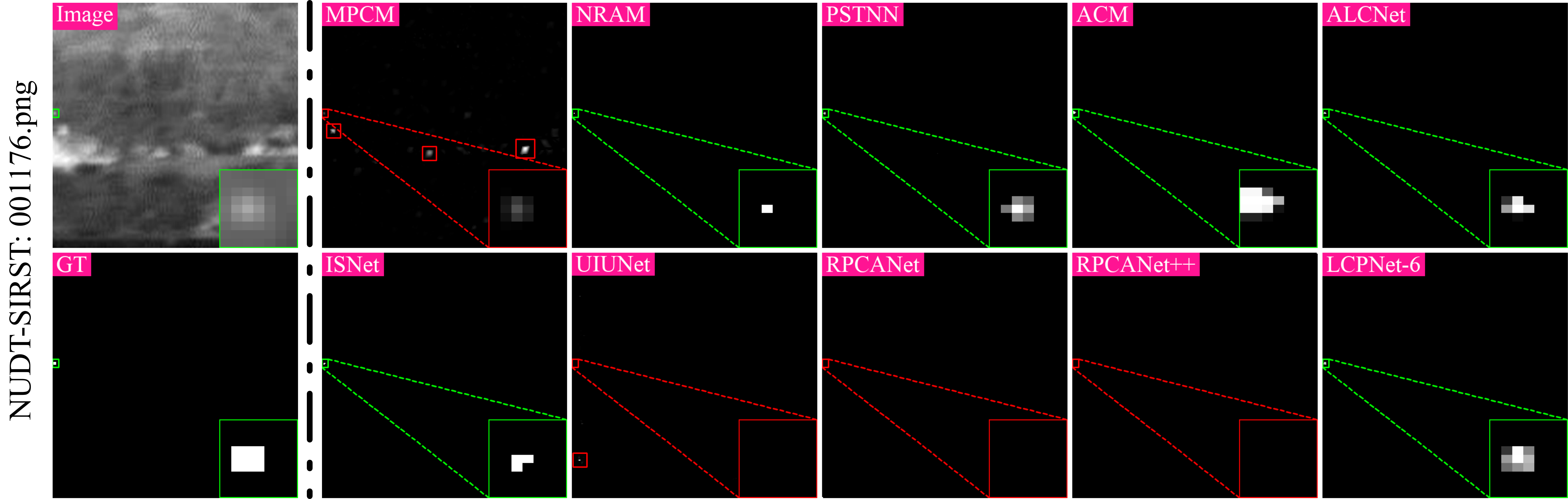}
  \caption{Qualitative comparison on image \textit{001176}. Green boxes denote correct detections, while red boxes denote false alarms or missed targets. Best view in color.}
  \label{fig:nudt-sirst-001176}
\end{figure*}

This is the same expansion as Eq.~\ref{eq:optimization_state_expansion}. In this expression, the contribution of each candidate state $\tilde{h}_t$ is controlled by its update gate $z_t$ and the subsequent retention gates $\prod_{i=t+1}^{k}(1-z_i)$. Therefore, $h_k$ is not determined only by $h_{k-1}$ or a fixed short window of previous states. Instead, it contains a gated sum of all candidate states $\{\tilde{h}_1,\tilde{h}_2,\ldots,\tilde{h}_k\}$ together with the retained initial state $h_0$. Thus, SOM has an unbounded effective memory order in its recurrent formulation, while the actual contribution of each historical state is adaptively controlled by the update gates.

\section{More Experimental Results}
\label{app:more_results}

This section provides additional qualitative results to complement the visual comparisons in the main paper. 
Fig.~\ref{fig:low-rank-2} further visualizes the low-rank property of the latent representation, where the latent features exhibit distinct low-rank characteristics in both visual and rank analysis.
Fig.~\ref{fig:heatmap_2} further visualizes the last-layer feature responses, where LCPNet shows more concentrated target-related activations and weaker background interference in challenging scenes. Figs.~\ref{fig:irstd-1k-xdu709}--\ref{fig:nudt-sirst-001176} present additional detection examples with different target appearances and background structures. Across these cases, LCPNet maintains compact responses around true targets and produces fewer clutter-induced false alarms, further supporting the robustness of the proposed latent-state unfolding design.

%% file: ref.bib
@inproceedings{tom1993morphology,
  title={Morphology-based algorithm for point target detection in infrared backgrounds},
  author={Tom, Victor T and Peli, Tamar and Leung, May and Bondaryk, Joseph E},
  booktitle={Signal and Data Processing of Small Targets 1993},
  volume={1954},
  pages={2--11},
  year={1993},
  organization={International Society for Optics and Photonics}
}

@inproceedings{deshpande1999max,
  title={Max-mean and max-median filters for detection of small targets},
  author={Deshpande, Suyog D and Er, Meng Hwa and Venkateswarlu, Ronda and Chan, Philip},
  booktitle={Signal and Data Processing of Small Targets 1999},
  volume={3809},
  pages={74--83},
  year={1999},
  organization={International Society for Optics and Photonics}
}

@article{chen2013local,
  title={A local contrast method for small infrared target detection},
  author={Chen, CL Philip and Li, Hong and Wei, Yantao and Xia, Tian and Tang, Yuan Yan},
  journal={IEEE Transactions on Geoscience and Remote Sensing},
  volume={52},
  number={1},
  pages={574--581},
  year={2013},
  publisher={IEEE}
}

@article{wei2016multiscale,
  title={Multiscale patch-based contrast measure for small infrared target detection},
  author={Wei, Yantao and You, Xinge and Li, Hong},
  journal={Pattern Recognition},
  volume={58},
  pages={216--226},
  year={2016},
  publisher={Elsevier}
}

@article{gao2013infrared,
  title={Infrared patch-image model for small target detection in a single image},
  author={Gao, Chenqiang and Meng, Deyu and Yang, Yi and Wang, Yongtao and Zhou, Xiaofang and Hauptmann, Alexander G},
  journal={IEEE Transactions on Image Processing},
  volume={22},
  number={12},
  pages={4996--5009},
  year={2013},
  publisher={IEEE}
}

@article{zhang2018infrared,
  title={Infrared small target detection via non-convex rank approximation minimization joint l 2, 1 norm},
  author={Zhang, Landan and Peng, Lingbing and Zhang, Tianfang and Cao, Siying and Peng, Zhenming},
  journal={Remote Sensing},
  volume={10},
  number={11},
  pages={1821},
  year={2018},
  publisher={MDPI}
}

@article{dai2017reweighted,
  title={Reweighted infrared patch-tensor model with both nonlocal and local priors for single-frame small target detection},
  author={Dai, Yimian and Wu, Yiquan},
  journal={IEEE journal of selected topics in applied earth observations and remote sensing},
  volume={10},
  number={8},
  pages={3752--3767},
  year={2017},
  publisher={IEEE}
}

@article{zhang2021infrared,
  title={Infrared small target detection via self-regularized weighted sparse model},
  author={Zhang, Tianfang and Peng, Zhenming and Wu, Hao and He, Yanmin and Li, Chaohai and Yang, Chunping},
  journal={Neurocomputing},
  volume={420},
  pages={124--148},
  year={2021},
  publisher={Elsevier}
}

@inproceedings{wang2019miss,
  title={Miss detection vs. false alarm: Adversarial learning for small object segmentation in infrared images},
  author={Wang, Huan and Zhou, Luping and Wang, Lei},
  booktitle={Proceedings of the IEEE/CVF International Conference on Computer Vision},
  pages={8509--8518},
  year={2019}
}

@inproceedings{dai2021asymmetric,
  title={Asymmetric Contextual Modulation for Infrared Small Target Detection},
  author={\vspace{0mm}Dai, Yimian and Wu, Yiquan and Zhou, Fei and Barnard, Kobus},
  booktitle={Proceedings of the IEEE/CVF Winter Conference on Applications of Computer Vision},
  pages={950--959},
  year={2021}
}

@article{zhang2023attention,
  title={Attention-guided pyramid context networks for detecting infrared small target under complex background},
  author={Zhang, Tianfang and Li, Lei and Cao, Siying and Pu, Tian and Peng, Zhenming},
  journal={IEEE Transactions on Aerospace and Electronic Systems},
  volume={59},
  number={4},
  pages={4250--4261},
  year={2023},
  publisher={IEEE}
}

@article{li2022dense,
  title={Dense nested attention network for infrared small target detection},
  author={Li, Boyang and Xiao, Chao and Wang, Longguang and Wang, Yingqian and Lin, Zaiping and Li, Miao and An, Wei and Guo, Yulan},
  journal={IEEE Transactions on Image Processing},
  volume={32},
  pages={1745--1758},
  year={2022},
  publisher={IEEE}
}

@article{wu2022uiu,
  title={UIU-Net: U-Net in U-Net for infrared small object detection},
  author={Wu, Xin and Hong, Danfeng and Chanussot, Jocelyn},
  journal={IEEE Transactions on Image Processing},
  volume={32},
  pages={364--376},
  year={2022},
  publisher={IEEE}
}

@inproceedings{liu2024infrared,
  title={Infrared small target detection with scale and location sensitivity},
  author={Liu, Qiankun and Liu, Rui and Zheng, Bolun and Wang, Hongkui and Fu, Ying},
  booktitle={Proceedings of the IEEE/CVF Conference on Computer Vision and Pattern Recognition},
  pages={17490--17499},
  year={2024}
}

@article{dai2021attentional,
  title={Attentional local contrast networks for infrared small target detection},
  author={Dai, Yimian and Wu, Yiquan and Zhou, Fei and Barnard, Kobus},
  journal={IEEE transactions on geoscience and remote sensing},
  volume={59},
  number={11},
  pages={9813--9824},
  year={2021},
  publisher={IEEE}
}

@inproceedings{zhang2022isnet,
  title={ISNet: Shape matters for infrared small target detection},
  author={Zhang, Mingjin and Zhang, Rui and Yang, Yuxiang and Bai, Haichen and Zhang, Jing and Guo, Jie},
  booktitle={Proceedings of the IEEE/CVF conference on computer vision and pattern recognition},
  pages={877--886},
  year={2022}
}

@inproceedings{zhang2024irsam,
  title={IRSAM: Advancing segment anything model for infrared small target detection},
  author={Zhang, Mingjin and Wang, Yuchun and Guo, Jie and Li, Yunsong and Gao, Xinbo and Zhang, Jing},
  booktitle={European Conference on Computer Vision},
  pages={233--249},
  year={2024},
  organization={Springer}
}

@article{wu2024saliency,
  title={Saliency at the helm: Steering infrared small target detection with learnable kernels},
  author={Wu, Fengyi and Liu, Anran and Zhang, Tianfang and Zhang, Luping and Luo, Junhai and Peng, Zhenming},
  journal={IEEE Transactions on Geoscience and Remote Sensing},
  volume={63},
  pages={1--14},
  year={2024},
  publisher={IEEE}
}

@inproceedings{yang2025pinwheel,
  title={Pinwheel-shaped convolution and scale-based dynamic loss for infrared small target detection},
  author={Yang, Jiangnan and Liu, Shuangli and Wu, Jingjun and Su, Xinyu and Hai, Nan and Huang, Xueli},
  booktitle={Proceedings of the AAAI Conference on Artificial Intelligence},
  volume={39},
  number={9},
  pages={9202--9210},
  year={2025}
}

@article{li2025ilnet,
  title={ILNet: Low-level matters for salient infrared small target detection},
  author={Li, Haoqing and Yang, Jinfu and Wang, Runshi and Xu, Yifei},
  journal={IEEE Transactions on Aerospace and Electronic Systems},
  year={2025},
  publisher={IEEE}
}

@inproceedings{wu2024rpcanet,
  title={RPCANet: Deep unfolding RPCA based infrared small target detection},
  author={Wu, Fengyi and Zhang, Tianfang and Li, Lei and Huang, Yian and Peng, Zhenming},
  booktitle={Proceedings of the IEEE/CVF Winter Conference on Applications of Computer Vision},
  pages={4809--4818},
  year={2024}
}

@article{wu2025rpcanet++,
  title={RPCANet++: Deep interpretable robust PCA for sparse object segmentation},
  author={Wu, Fengyi and Dai, Yimian and Zhang, Tianfang and Ding, Yixuan and Yang, Jian and Cheng, Ming-Ming and Peng, Zhenming},
  journal={arXiv preprint arXiv:2508.04190},
  year={2025}
}

@article{boyd2011distributed,
  title={Distributed optimization and statistical learning via the alternating direction method of multipliers},
  author={Boyd, Stephen and Parikh, Neal and Chu, Eric and Peleato, Borja and Eckstein, Jonathan},
  journal={Foundations and Trends in Machine Learning},
  volume={3},
  number={1},
  pages={1--122},
  year={2011},
  doi={10.1561/2200000016}
}

@book{nesterov2004introductory,
  title={Introductory Lectures on Convex Optimization: A Basic Course},
  author={Nesterov, Yurii},
  series={Applied Optimization},
  volume={87},
  year={2004},
  publisher={Springer},
  doi={10.1007/978-1-4419-8853-9}
}

@inproceedings{cho2014learning,
  title={Learning Phrase Representations using RNN Encoder--Decoder for Statistical Machine Translation},
  author={Cho, Kyunghyun and van Merrienboer, Bart and Gulcehre, Caglar and Bahdanau, Dzmitry and Bougares, Fethi and Schwenk, Holger and Bengio, Yoshua},
  booktitle={Proceedings of the 2014 Conference on Empirical Methods in Natural Language Processing},
  pages={1724--1734},
  year={2014},
  publisher={Association for Computational Linguistics},
  doi={10.3115/v1/D14-1179}
}

@inproceedings{cisse2017parseval,
  title={Parseval networks: Improving robustness to adversarial examples},
  author={Cisse, Moustapha and Bojanowski, Piotr and Grave, Edouard and Dauphin, Yann and Usunier, Nicolas},
  booktitle={International conference on machine learning},
  pages={854--863},
  year={2017},
  organization={PMLR}
}

@article{miyato2018spectral,
  title={Spectral normalization for generative adversarial networks},
  author={Miyato, Takeru and Kataoka, Toshiki and Koyama, Masanori and Yoshida, Yuichi},
  journal={arXiv preprint arXiv:1802.05957},
  year={2018}
}

@inproceedings{ioffe2015batch,
  title={Batch normalization: Accelerating deep network training by reducing internal covariate shift},
  author={Ioffe, Sergey and Szegedy, Christian},
  booktitle={International conference on machine learning},
  pages={448--456},
  year={2015},
  organization={pmlr}
}

@inproceedings{wu2018group,
  title={Group normalization},
  author={Wu, Yuxin and He, Kaiming},
  booktitle={Proceedings of the European conference on computer vision (ECCV)},
  pages={3--19},
  year={2018}
}

@article{zhao2022single,
  title={Single-frame infrared small-target detection: A survey},
  author={Zhao, Mingjing and Li, Wei and Li, Lu and Hu, Jin and Ma, Pengge and Tao, Ran},
  journal={IEEE Geoscience and Remote Sensing Magazine},
  volume={10},
  number={2},
  pages={87--119},
  year={2022},
  publisher={IEEE},
  doi={10.1109/MGRS.2022.3145502}
}

@article{kou2023infrared,
  title={Infrared small target segmentation networks: A survey},
  author={Kou, Renke and Wang, Chunping and Peng, Zhenming and Zhao, Zhihe and Chen, Yaohong and Han, Jinhui and Huang, Fuyu and Yu, Ying and Fu, Qiang},
  journal={Pattern Recognition},
  volume={143},
  pages={109788},
  year={2023},
  publisher={Elsevier},
  doi={10.1016/j.patcog.2023.109788}
}

@article{candes2011robust,
  title={Robust principal component analysis?},
  author={Cand{\`e}s, Emmanuel J. and Li, Xiaodong and Ma, Yi and Wright, John},
  journal={Journal of the ACM},
  volume={58},
  number={3},
  pages={1--37},
  year={2011},
  publisher={Association for Computing Machinery},
  doi={10.1145/1970392.1970395}
}

@article{monga2021algorithm,
  title={Algorithm unrolling: Interpretable, efficient deep learning for signal and image processing},
  author={Monga, Vishal and Li, Yuelong and Eldar, Yonina C.},
  journal={IEEE Signal Processing Magazine},
  volume={38},
  number={2},
  pages={18--44},
  year={2021},
  publisher={IEEE},
  doi={10.1109/MSP.2020.3016905}
}

@inproceedings{shi2015convolutional,
  title={Convolutional LSTM network: A machine learning approach for precipitation nowcasting},
  author={Shi, Xingjian and Chen, Zhourong and Wang, Hao and Yeung, Dit-Yan and Wong, Wai-Kin and Woo, Wang-Chun},
  booktitle={Advances in Neural Information Processing Systems},
  volume={28},
  pages={802--810},
  year={2015},
}

@inproceedings{gregor2010learning,
  title={Learning fast approximations of sparse coding},
  author={Gregor, Karol and LeCun, Yann},
  booktitle={Proceedings of the 27th International Conference on Machine Learning},
  pages={399--406},
  year={2010},
  publisher={Omnipress},
}

@inproceedings{yang2016deep,
  title={Deep ADMM-Net for compressive sensing MRI},
  author={Yang, Yan and Sun, Jian and Li, Huibin and Xu, Zongben},
  booktitle={Advances in Neural Information Processing Systems},
  volume={29},
  year={2016},
  publisher={Curran Associates, Inc.},
}

@inproceedings{zhang2018ista,
  title={ISTA-Net: Interpretable optimization-inspired deep network for image compressive sensing},
  author={Zhang, Jian and Ghanem, Bernard},
  booktitle={2018 IEEE/CVF Conference on Computer Vision and Pattern Recognition},
  pages={1828--1837},
  year={2018},
  publisher={IEEE},
  doi={10.1109/CVPR.2018.00196}
}

@article{aggarwal2018modl,
  title={MoDL: Model-based deep learning architecture for inverse problems},
  author={Aggarwal, Hemant Kumar and Mani, Merry P. and Jacob, Mathews},
  journal={IEEE Transactions on Medical Imaging},
  volume={38},
  number={2},
  pages={394--405},
  year={2018},
  publisher={IEEE},
  doi={10.1109/TMI.2018.2865356}
}

@inproceedings{cai2021learned,
  title={Learned robust PCA: A scalable deep unfolding approach for high-dimensional outlier detection},
  author={Cai, HanQin and Liu, Jialin and Yin, Wotao},
  booktitle={Advances in Neural Information Processing Systems},
  volume={34},
  pages={16977--16989},
  year={2021},
  publisher={Curran Associates, Inc.},
}

@inproceedings{ho2020denoising,
  title={Denoising diffusion probabilistic models},
  author={Ho, Jonathan and Jain, Ajay and Abbeel, Pieter},
  booktitle={Advances in Neural Information Processing Systems},
  volume={33},
  pages={6840--6851},
  year={2020},
  publisher={Curran Associates, Inc.},
}

@article{hochreiter1997long,
  title={Long Short-Term Memory},
  author={Hochreiter, Sepp and Schmidhuber, J{\"u}rgen},
  journal={Neural Computation},
  volume={9},
  number={8},
  pages={1735--1780},
  year={1997},
  publisher={MIT Press},
  doi={10.1162/neco.1997.9.8.1735}
}

@misc{putzky2017recurrent,
  title={Recurrent Inference Machines for Solving Inverse Problems},
  author={Putzky, Patrick and Welling, Max},
  year={2017},
  eprint={1706.04008},
  archivePrefix={arXiv},
  primaryClass={cs.NE},
}

@article{adler2018learned,
  title={Learned Primal-Dual Reconstruction},
  author={Adler, Jonas and {\"O}ktem, Ozan},
  journal={IEEE Transactions on Medical Imaging},
  volume={37},
  number={6},
  pages={1322--1332},
  year={2018},
  publisher={IEEE},
  doi={10.1109/TMI.2018.2799231}
}

@article{hosseini2020dense,
  title={Dense Recurrent Neural Networks for Accelerated {MRI}: History-Cognizant Unrolling of Optimization Algorithms},
  author={Hosseini, Seyed Amir Hossein and Yaman, Burhaneddin and Moeller, Steen and Hong, Mingyi and Akcakaya, Mehmet},
  journal={IEEE Journal of Selected Topics in Signal Processing},
  volume={14},
  number={6},
  pages={1280--1291},
  year={2020},
  publisher={IEEE},
  doi={10.1109/JSTSP.2020.3003170}
}

@inproceedings{song2021memory,
  title={Memory-Augmented Deep Unfolding Network for Compressive Sensing},
  author={Song, Jiechong and Chen, Bin and Zhang, Jian},
  booktitle={Proceedings of the 29th ACM International Conference on Multimedia},
  pages={4249--4258},
  year={2021},
  publisher={ACM},
  doi={10.1145/3474085.3475562}
}

@inproceedings{yiasemis2022recurrent,
  title={Recurrent Variational Network: A Deep Learning Inverse Problem Solver Applied to the Task of Accelerated {MRI} Reconstruction},
  author={Yiasemis, George and Sonke, Jan-Jakob and Sanchez, Clarisa and Teuwen, Jonas},
  booktitle={2022 IEEE/CVF Conference on Computer Vision and Pattern Recognition},
  pages={722--731},
  year={2022},
  publisher={IEEE},
  doi={10.1109/CVPR52688.2022.00081}
}

@misc{gu2022s4,
  title={Efficiently Modeling Long Sequences with Structured State Spaces},
  author={Gu, Albert and Goel, Karan and R{\'e}, Christopher},
  year={2022},
  eprint={2111.00396},
  archivePrefix={arXiv},
  primaryClass={cs.LG},
}

@misc{gu2024mamba,
  title={Mamba: Linear-Time Sequence Modeling with Selective State Spaces},
  author={Gu, Albert and Dao, Tri},
  year={2024},
  eprint={2312.00752},
  archivePrefix={arXiv},
  primaryClass={cs.LG},
}

@misc{zhu2024visionmamba,
  title={Vision Mamba: Efficient Visual Representation Learning with Bidirectional State Space Model},
  author={Zhu, Lianghui and Liao, Bencheng and Zhang, Qian and Wang, Xinlong and Liu, Wenyu and Wang, Xinggang},
  year={2024},
  eprint={2401.09417},
  archivePrefix={arXiv},
  primaryClass={cs.CV},
}

@misc{liu2024vmamba,
  title={VMamba: Visual State Space Model},
  author={Liu, Yue and Tian, Yunjie and Zhao, Yuzhong and Yu, Hongtian and Xie, Lingxi and Wang, Yaowei and Ye, Qixiang and Jiao, Jianbin and Liu, Yunfan},
  year={2024},
  eprint={2401.10166},
  archivePrefix={arXiv},
  primaryClass={cs.CV},
}

@misc{metzler2017learned,
  title={Learned {D-AMP}: Principled Neural Network Based Compressive Image Recovery},
  author={Metzler, Christopher A. and Mousavi, Ali and Baraniuk, Richard G.},
  year={2017},
  eprint={1704.06625},
  archivePrefix={arXiv},
  primaryClass={stat.ML},
}

@article{zhang2021ampnet,
  title={{AMP-Net}: Denoising Based Deep Unfolding for Compressive Image Sensing},
  author={Zhang, Zhonghao and Liu, Yipeng and Liu, Jiani and Wen, Fei and Zhu, Ce},
  journal={IEEE Transactions on Image Processing},
  volume={30},
  pages={1487--1500},
  year={2021},
  publisher={IEEE},
  doi={10.1109/TIP.2020.3044472}
}

@article{hammernik2017learning,
  title={Learning a Variational Network for Reconstruction of Accelerated {MRI} Data},
  author={Hammernik, Kerstin and Klatzer, Teresa and Kobler, Erich and Recht, Michael P. and Sodickson, Daniel K. and Pock, Thomas and Knoll, Florian},
  journal={Magnetic Resonance in Medicine},
  volume={79},
  number={6},
  pages={3055--3071},
  year={2017},
  publisher={Wiley},
  doi={10.1002/mrm.26977}
}

@incollection{sriram2020end,
  title={End-to-End Variational Networks for Accelerated {MRI} Reconstruction},
  author={Sriram, Anuroop and Zbontar, Jure and Murrell, Tullie and Defazio, Aaron and Zitnick, C. Lawrence and Yakubova, Nafissa and Knoll, Florian and Johnson, Patricia},
  booktitle={Medical Image Computing and Computer Assisted Intervention -- MICCAI 2020},
  pages={64--73},
  year={2020},
  publisher={Springer International Publishing},
  doi={10.1007/978-3-030-59713-9_7}
}

@misc{sohldickstein2015deep,
  title={Deep Unsupervised Learning Using Nonequilibrium Thermodynamics},
  author={Sohl-Dickstein, Jascha and Weiss, Eric A. and Maheswaranathan, Niru and Ganguli, Surya},
  year={2015},
  eprint={1503.03585},
  archivePrefix={arXiv},
  primaryClass={cs.LG},
}

@misc{song2019generative,
  title={Generative Modeling by Estimating Gradients of the Data Distribution},
  author={Song, Yang and Ermon, Stefano},
  year={2020},
  eprint={1907.05600},
  archivePrefix={arXiv},
  primaryClass={cs.LG},
}

@misc{song2020denoising,
  title={Denoising Diffusion Implicit Models},
  author={Song, Jiaming and Meng, Chenlin and Ermon, Stefano},
  year={2022},
  eprint={2010.02502},
  archivePrefix={arXiv},
  primaryClass={cs.LG},
}

@misc{nichol2021improved,
  title={Improved Denoising Diffusion Probabilistic Models},
  author={Nichol, Alex and Dhariwal, Prafulla},
  year={2021},
  eprint={2102.09672},
  archivePrefix={arXiv},
  primaryClass={cs.LG},
}

@article{peng2019infrared,
  title={Infrared small-target detection based on multi-directional multi-scale high-boost response},
  author={Peng, Lingbing and Zhang, Tianfang and Huang, Suqi and Pu, Tian and Liu, Yuhan and Lv, Yuxiao and Zheng, Yunchang and Peng, Zhenming},
  journal={Optical Review},
  volume={26},
  number={6},
  pages={568--582},
  year={2019},
  publisher={Springer}
}

@article{zhang2019infrared,
  title={Infrared small target detection based on non-convex optimization with Lp-norm constraint},
  author={Zhang, Tianfang and Wu, Hao and Liu, Yuhan and Peng, Lingbing and Yang, Chunping and Peng, Zhenming},
  journal={Remote Sensing},
  volume={11},
  number={5},
  pages={559},
  year={2019},
  publisher={MDPI}
}

@article{yuan2024sctransnet,
  title={SCTransNet: Spatial-channel cross transformer network for infrared small target detection},
  author={Yuan, Shuai and Qin, Hanlin and Yan, Xiang and Akhtar, Naveed and Mian, Ajmal},
  journal={IEEE Transactions on Geoscience and Remote Sensing},
  volume={62},
  pages={1--15},
  year={2024},
  publisher={IEEE}
}

@article{xiong2025drpca,
  title={DRPCA-Net: Make robust PCA great again for infrared small target detection},
  author={Xiong, Zihao and Zhou, Fei and Wu, Fengyi and Yuan, Shuai and Fu, Maixia and Peng, Zhenming and Yang, Jian and Dai, Yimian},
  journal={IEEE Transactions on Geoscience and Remote Sensing},
  year={2025},
  publisher={IEEE}
}

@article{zhang2023optimization,
  title={Optimization-inspired Cumulative Transmission Network for image compressive sensing},
  author={Zhang, Tianfang and Li, Lei and Peng, Zhenming},
  journal={Knowledge-Based Systems},
  volume={279},
  pages={110963},
  year={2023},
  publisher={Elsevier}
}

@article{yuan2026sp,
  title={SP-KAN: Sparse-sine perception Kolmogorov--Arnold networks for infrared small target detection},
  author={Yuan, Shuai and Liu, Yu and Zhang, Xiaopei and Yan, Xiang and Qin, Hanlin and Akhtar, Naveed},
  journal={ISPRS Journal of Photogrammetry and Remote Sensing},
  volume={234},
  pages={1--19},
  year={2026},
  publisher={Elsevier}
}
